\documentclass[runningheads]{llncs}
\usepackage{graphicx}
\usepackage{amsmath,amssymb} 
\usepackage{color}
\usepackage{amsmath}
\usepackage{amssymb}
\usepackage{multirow}
\usepackage{makecell}
\usepackage{subfigure}
\usepackage[colorlinks,linkcolor=red]{hyperref}

\usepackage[width=122mm,left=12mm,paperwidth=146mm,height=193mm,top=12mm,paperheight=217mm]{geometry}
\begin{document}
\pagestyle{headings}
\mainmatter
\def\ECCV18SubNumber{2046}  

\title{Deep Texture and Structure Aware Filtering Network for Image Smoothing} 



\author{Kaiyue Lu, Shaodi You, Nick Barnes}

\institute{Research School of Engineering, Australian National University\\ Data61, CSIRO\\\tt\small \{Kaiyue.Lu, Shaodi.You, Nick.Barnes\}@data61.csiro.au}

\maketitle
\begin{abstract}
Image smoothing is a fundamental task in computer vision, that aims to retain salient structures and remove insignificant textures. In this paper, we aim to address the fundamental shortcomings of existing image smoothing methods, which cannot properly distinguish textures and structures with similar low-level appearance. 
While deep learning approaches have started to explore the preservation of structure through image smoothing, existing work does not yet properly address textures. To this end, we generate a large dataset by blending natural textures with clean structure-only images, and then build a texture prediction network (TPN) that predicts the location and magnitude of textures. We then combine the TPN with a semantic structure prediction network (SPN) so that the final texture and structure aware filtering network (TSAFN) is able to identify the textures to remove (``texture-awareness") and the structures to preserve (``structure-awareness"). The proposed model is easy to understand and implement, and shows excellent performance on real images in the wild as well as our generated dataset.
\end{abstract}


\section{Introduction}

Image smoothing, a fundamental technology in image processing and computer vision, aims to clean images by retaining salient structures (to the \textbf{\textit{structure-only image}}) and removing insignificant textures (to the \textbf{\textit{texture-only image}}), with various applications including denoising \cite{32gu2014weighted}, detail enhancement \cite{31fattal2007multiscale}, image abstraction \cite{30winnemoller2006real} and segmentation \cite{61wang2012image}. 

There are mainly two types of methods for image smoothing: (1) kernel-based methods, that calculate the average of the neighborhood for texture pixels while trying to retain the original value for structural pixels, such as the guided filter (GF) \cite{2he2013guided}, rolling guidance filter (RGF) \cite{11zhang2014rolling}, segment graph filter (SGF) \cite{3zhang2015segment} and so on; and (2) separation-based methods, which decompose the image into a structure layer and a texture layer, such as relative total variation (RTV) \cite{23xu2012structure}, fast L0 \cite{26nguyen2015fast}, and static and dynamic guidance filter (SDF) \cite{12ham2015robust,13ham2017robust}.
Traditional approaches rely on hand-crafted features and/or prior knowledge to distinguish textures from structures. These features are based entirely on low-level appearance, and generally assume that structures always have larger gradients, and textures are just smaller oscillations in color intensities.

\begin{figure}[!t]
\centering

\subfigure{\includegraphics[width=\linewidth]{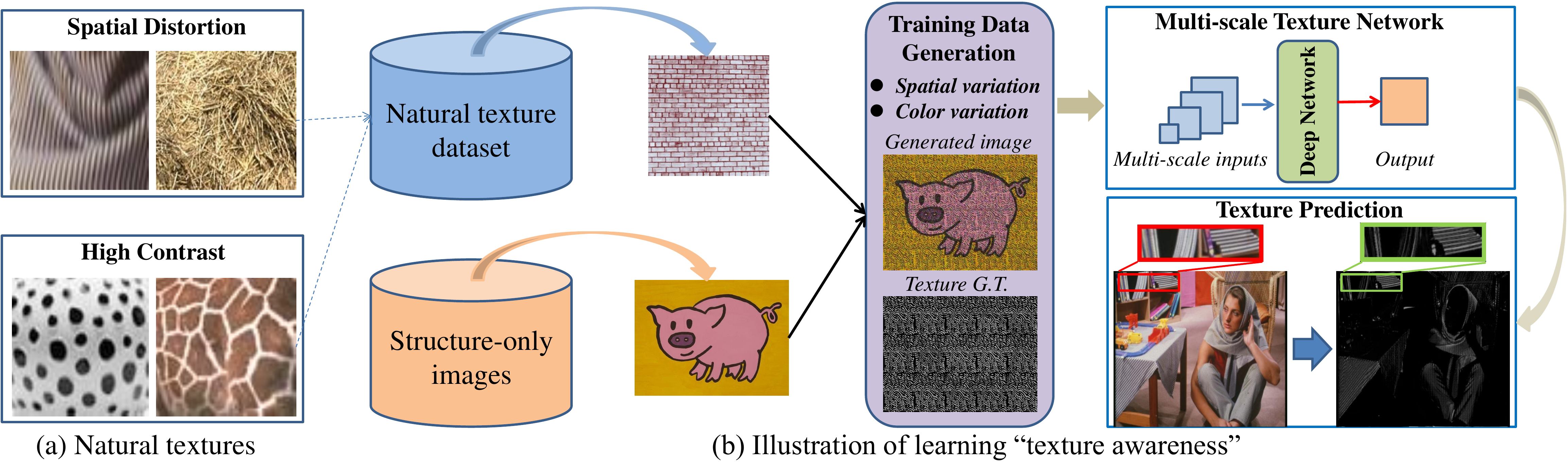}}\\

\subfigure{\includegraphics[width=\linewidth]{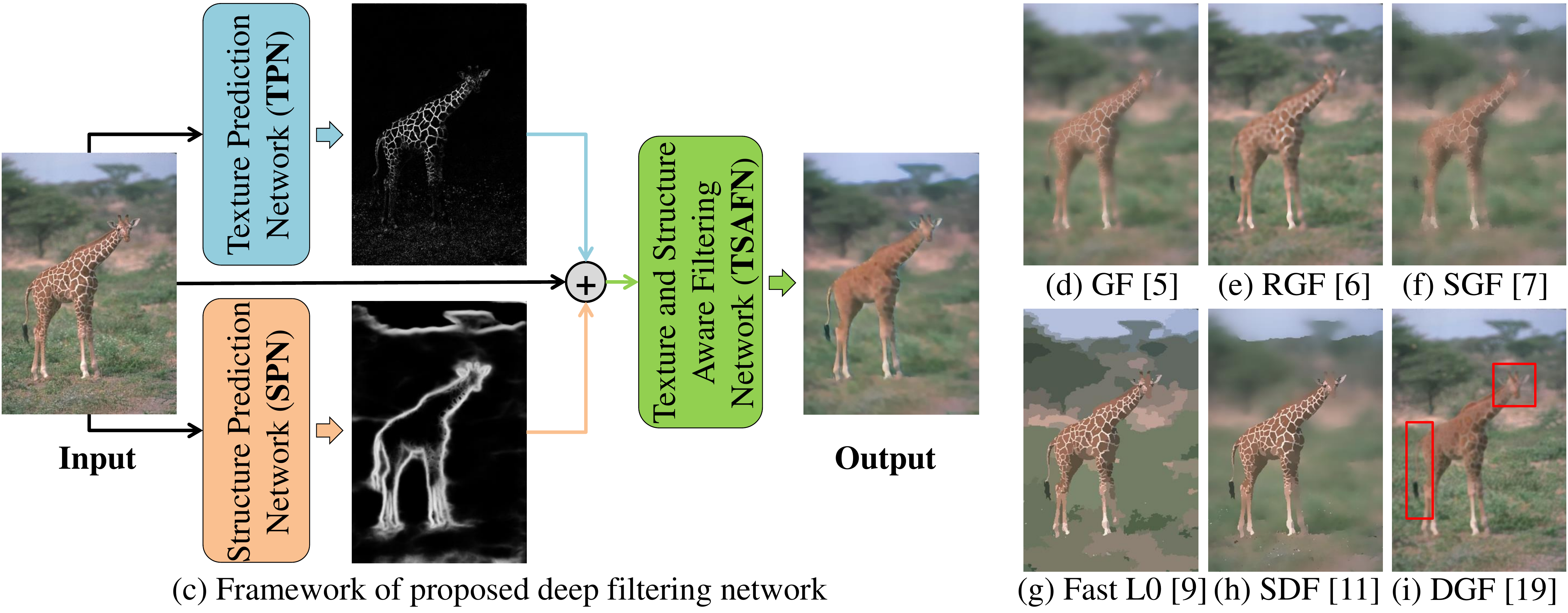}}\\

\vspace{-9pt}

\caption{(a) Texture in natural images is often hard to identify due to spatial distortion and high contrast. (b) Illustration of learning ``texture awareness". We generate training data by adding spatial and color variations to natural texture patterns and blending them with structure-only images, and then use the result to train a multi-scale texture network with texture ground-truth. We test the network on both generated data and natural images. (c) Our proposed deep filtering network is composed of a texture prediction network (TPN) for predicting textures (white stripes with high-contrast); a structure prediction network (SPN) for extracting structures (the giraffe's boundary, which has relatively low contrast to the background); and a texture and structure aware filtering network (TSAFN) for image smoothing.
(d)-(i) Existing methods cannot distinguish low-contrast structures from high-contrast textures effectively. 
}
\label{fig:fig1}
\vspace{-15pt}
\end{figure}

In fact, it is quite difficult to identify textures. The main reasons are twofold: (1) textures are essentially repeated patterns regularly or irregularly distributed within object structures, and they may show significant spatial distortions in an image (as shown in Fig.~\ref{fig:fig1}(a)), making it impossible to fully define them mathematically; (2) in some images there are strong textures with large gradients and color contrast to the background, which are easy to confuse with structures (such as the white stripes on the giraffe's body in Fig.~\ref{fig:fig1}(c)). We see from Fig.~\ref{fig:fig1} that GF, RGF, SGF, fast L0, and SDF perform poorly on the giraffe image. The textures are either not removed at all, or suppressed with the structure severely blurred. This is because the hand-crafted nature of these filters makes them less robust when applied to various types of textures, and also leads to poor discrimination of textures and structures. Some other methods \cite{46xu2015deep,47liu2016learning,48li2016deep,49Fan_2017_ICCV,52Chen_2017_ICCV,50DBLP:journals/corr/FanWHC17,51DBLP:journals/corr/ShenCTJ17} take advantage of deep neural networks, and aim for better performance by extracting richer information. However, existing networks use the output of various hand-crafted filters as ground-truth during training. These deep learning approaches are thus limited by the shortcomings of hand-crafted filters, and cannot learn how to effectively distinguish textures from structures.

A recently-proposed double-guided filter (DGF) \cite{53Lu_2017_DICTA} addresses this issue by introducing the idea of ``texture guidance", which  infers the location of texture, and combines it with ``structure guidance" to achieve both goals of texture removal and structure preservation. However, DGF uses a hand-crafted separation-based algorithm called Structure Gradient and Texture Decorrelating (SGTD) \cite{9liu2013sgtd} to construct the texture confidence map that still cannot essentially overcome the natural deficiency. We argue that this is not true ``texture awareness", because in many cases, some structures are inevitably blurred when the filter tries to remove strong textures after several iterations. As can be seen in Fig.~\ref{fig:fig1}(i), although the stripe textures are largely smoothed out, the structure of the giraffe is unexpectedly blurred, especially around the head and the tail (red boxes).

In this paper, we hold the idea that ``texture awareness" should reflect both the \textit{texture region} (where the texture is) and \textit{texture magnitude} (texture with high contrast to the background is harder to remove). Thus, we take advantage of deep learning and propose a texture prediction network (TPN) that aims to learn textures from natural images. However, since there are no available datasets containing natural images with labeled texture regions, we make use of texture-only datasets \cite{55cimpoi14describing,56dana1999reflectance}. The process of learning ``texture awareness" is shown in Fig.~\ref{fig:fig1}(b). Specifically, we generate the training data by adding spatial and color variations to natural texture patterns and blending them with the structure-only image. Then we construct a multi-scale network (containing different levels of contextual information) to train these images with texture ground-truth (G.T. in short). The proposed TRN is able to predict textures through a full consideration of both high-level statistics, \textit{e.g.}, repetition, tiling, spatial varying distortion; and low-level appearance, \textit{e.g.}, gradient. The network achieves good performance on our generated testing data, and can also generalize well to natural images, effectively locating texture regions and measuring texture magnitude by assigning different confidences, as shown in Fig.~\ref{fig:fig1}(b). More details can be found in Section~\ref{sect:tpn}.

For the full problem, we are inspired by the idea of ``double guidance" introduced in \cite{53Lu_2017_DICTA} and propose a deep neural network based filter that learns to predict textures to remove (``texture-awareness" by our TPN) and structures to preserve (``structure-awareness" by HED semantic edge detection \cite{54xie2015holistically}). This is an end-to-end image smoothing architecture which we refer to as ``Texture and Structure Aware Filtering Network" (TSAFN), as shown in Fig.~\ref{fig:fig1}(c). The network is trained with our own generated dataset. Different from the work in \cite{53Lu_2017_DICTA}, we propose more effective methods for generating texture and structure guidance, and replace the hand-crafted kernel filter with a deep learning based one to achieve a more consistent and effective combination of these two types of guidance. Experimental results show that our proposed filter outperforms DGF \cite{53Lu_2017_DICTA} significantly in terms of both effectiveness and efficiency, achieves state-of-the-art performance on our dataset, and generalizes well to natural images.

The main contributions of this paper are: 
(1) To the best of our knowledge, we are the first to propose 
deep neural networks to robustly predict textures in natural images.
(2) We present a large dataset that enables training texture prediction and image smoothing.
(3) We propose an end-to-end deep neural network for image smoothing that achieves both ``structure-awareness" and ``texture-awareness", and outperforms existing methods on challenging natural images.

\vspace{-7pt}
\section{Related Work}


\paragraph{\textbf{Texture extraction from structures}}
\vspace{-6pt}
The basic assumption of this type of work is that an image can be decomposed into structure and texture layers (the structure layer is a smoothed version of the input and contains salient structures, while the texture layer contains insignificant details or textures). The pioneering work, Total Variation \cite{22rudin1992nonlinear}, aims to minimize the quadratic difference between the input and output images to maintain structure consistency with the gradient loss as an additional penalty. Later works retain the quadratic form and propose other regularizer terms or features (\textit{gradient loss is still necessary to keep the structures as sharp as possible}), such as weighted least squares (WLS) \cite{5farbman2008edge},  $\ell_0$ norm smoothing \cite{4xu2011image,26nguyen2015fast}, $\ell_1$ norm smoothing \cite{25bi20151}, local extrema \cite{28subr2009edge}, structure gradient and texture decorrelating (SGTD) \cite{9liu2013sgtd}. Other works also focuses on accelerating the optimization \cite{24buades2010fast} or improving existing algorithms \cite{57Liu_2017_ICCV}. There are two general issues that have not been handled effectively in existing work. Firstly, as they are largely dependent on gradient information, these methods \textit{lack discrimination of textures and structures}, especially when they have similar low-level appearance, particularly in terms of scale or magnitude. Secondly, all the objective functions are \textit{manually defined}, and may not be adaptive and robust to the huge variety of possible textures, especially in natural images.


\vspace{-6pt}
\paragraph{\textbf{Image smoothing with guidance}}
The guidance image can provide structure information to help repair and sharpen structures in the target image. Since adding guidance into separation-based methods may make it harder to optimize, this idea is more widely used in kernel-based methods. Static guidance refers to the use of a fixed guidance image, such as the bilateral filter \cite{1tomasi1998bilateral}, joint bilateral filter \cite{10petschnigg2004digital}, and guided filter \cite{2he2013guided}. To make the guidance more structure-aware, existing filters also employ techniques such as leverage tree distance \cite{15bao2014tree}, superpixels \cite{3zhang2015segment}, region covariances \cite{14karacan2013structure}, co-occurrence matrix \cite{34Jevnisek_2017_CVPR}, propagation distance \cite{16rick2015propagated}, multipoint estimation \cite{20tan2014multipoint}, fully connected regions \cite{21dai2015fully} and edge maps \cite{6yang2016semantic,18cho2014bilateral,29zang2015guided}. In contrast, dynamic guidance methods update the guidance image to suppress more details \cite{11zhang2014rolling,12ham2015robust,13ham2017robust} by iteratively refining the target image. Overall, the aforementioned guidance methods only address structure information, or assume that structures and textures can be sufficiently distinguished with a single guidance. However, in most cases, structures and textures interfere with each other severely. Lu \textit{et al.} \cite{53Lu_2017_DICTA} addresses this issue by introducing the concept of ``texture guidance", which infers texture regions by normalizing the texture layer separated by SGTD \cite{9liu2013sgtd} to construct the texture confidence map. They then naively combine it with structure guidance to form a double-guided kernel filter. However, this method is still largely dependent on hand-crafted features (in particular it relies on the hand-crafted SGTD to infer textures, which is not robust in essence). Structures may be blurred when the filter tries to smooth out strong textures after several iterations.
\vspace{-6pt}
\paragraph{\textbf{Deep image smoothing}}

Deep learning has been widely used in low-level vision tasks, such as super resolution \cite{58Tai_2017_CVPR}, deblurring \cite{60Nah_2017_CVPR} and dehazing \cite{59cai2016dehazenet}. Compared with non-learning approaches, deep learning is able to extract richer information from images. In image smoothing, current deep filtering models all focus on approximating and accelerating existing non-learned filters. \cite{46xu2015deep} is the pioneering paper, where the learning is performed on the gradient domain and the output is reconstructed from the refined gradients produced by the deep network. Liu \textit{et al.} \cite{47liu2016learning} take advantage of both convolutional networks (for perceiving salient structures) and recurrent networks (for producing smoothing output in a data-driven manner). Li \textit{et al.} \cite{48li2016deep} fuse the features from the original input and guidance image together and then produce the guided smoothing result (this work is mainly for upsampling). Fan \textit{et al.} \cite{49Fan_2017_ICCV} first construct a network called E-CNN to predict the edge/structure confidence map based on gradients, and then use it to guide the filtering network called I-CNN. Similar work can be found in \cite{50DBLP:journals/corr/FanWHC17} by the same authors. Most recent works mainly focus on extracting richer information from input images (\cite{51DBLP:journals/corr/ShenCTJ17} introduces a convolutional neural pyramid to extract features of different scales, and \cite{52Chen_2017_ICCV} utilizes context aggregation networks to include more contextual information) and yielding more satisfying results. One common issue is all of these approaches have to take the output of existing filters as ground-truth and cannot function as an independent filter. Their focus is limited to how similarly to the learned filter they can perform, and how fast it can accelerate computation. This deviates from the task of image smoothing itself. Moreover, since these methods aim to mimic existing filters, they are unable to overcome their deficiency in discriminating textures.

\vspace{-10pt}

\section{Texture Prediction}
\label{sect:tpn}
\vspace{-11pt}
In this section, we give insights on textures in natural images, which inspire the design of the texture prediction network (TPN) and the dataset for training.
\vspace{-15pt}

\subsection{What is texture?}
\begin{figure}[tb]
\centering
\includegraphics[width=\linewidth]{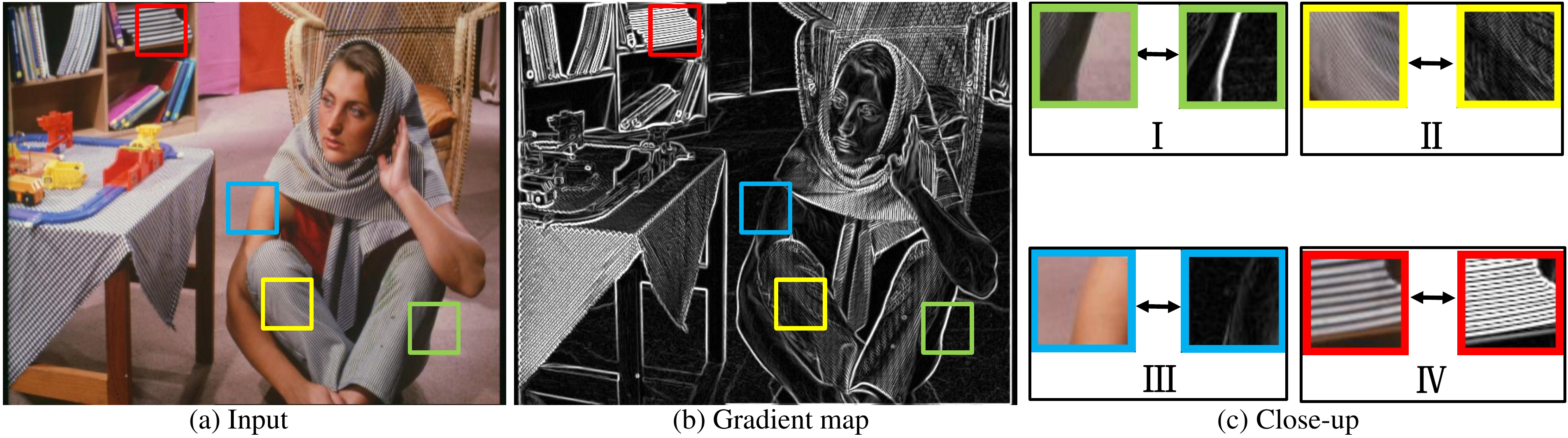}\\
\vspace{-12pt}
\caption{Close observation of structures and textures. In contrast with the assumptions used in existing methods, large gradients do not necessarily indicate structures (\uppercase\expandafter{\romannumeral4}), and small gradients may also belong to structures (\uppercase\expandafter{\romannumeral3}). The challenge to distinguish them motivates us to propose two independent texture and structure guidances.}
\label{fig:close}
\vspace{-15pt}
\end{figure}

\vspace{-8pt}
\paragraph{\textbf{Appearance of texture}}
It is well known that many different types of textures occur in nature and it is difficult to fully define them mathematically. Generally speaking, textures are repeated patterns regularly or irregularly distributed within object structures. For example, in Fig.~\ref{fig:fig1}(c), the white stripes on the giraffe's surface are recognized as textures. In Fig.~\ref{fig:close}, textures are widely spread in the image on clothes, books, and the table cloth. For cognition and vision tasks, an intuitive observation is that the removal of these textures will not affect the spatial structure of objects. Thus, they can be removed by image smoothing as a preprocessing step for other visual tasks. 
\vspace{-6pt}

\paragraph{\textbf{Textures do not necessarily have small gradients}}
Existing methods generally assume that textures are minor oscillations and have small gradients. Thus, they can easily hand-craft the filter or loss function. However, in many cases, textures may also have large gradients, \textit{e.g.}, the white stripes on the giraffe's body in Fig.~\ref{fig:fig1}(b), and the stripes occurring on the books in close-up \uppercase\expandafter{\romannumeral4} of Fig.~\ref{fig:close}(c). 
Therefore, defining textures purely based on local contrast is insufficient.

\vspace{-6pt}
\paragraph{\textbf{Mathematically modeling texture repetition is non-trivial}}
By definition, textures are patterns with spatial repetitions. However, modeling and describing the repetition is non-trivial due to the existence of various distortions (see Fig.~\ref{fig:fig1}(a)).

\vspace{-6pt}
\paragraph{\textbf{Learn to predict textures} }
To tackle these issues, we take advantage of deep neural networks. Provided sufficient training examples are available, the network is able to learn to predict textures without explicit modeling.

\vspace{-7pt}
\subsection{Dataset Generation}
\vspace{-4pt}
We aim to provide a dataset so that a deep network can learn to predict textures. Ideally, we would like to learn directly from natural images. However, manually annotating pixel-wise labels plus alpha-matting would be prohibitively costly. Moreover, it would require a full range of textures, each with a full range of distortions in a broad array of natural scenes. Therefore, we propose a strategy to generate the training and testing data. Later, we will demonstrate that the proposed network is able to predict textures in the wild successfully.

\begin{figure*}[tb]
\centering
\includegraphics[width=\linewidth]{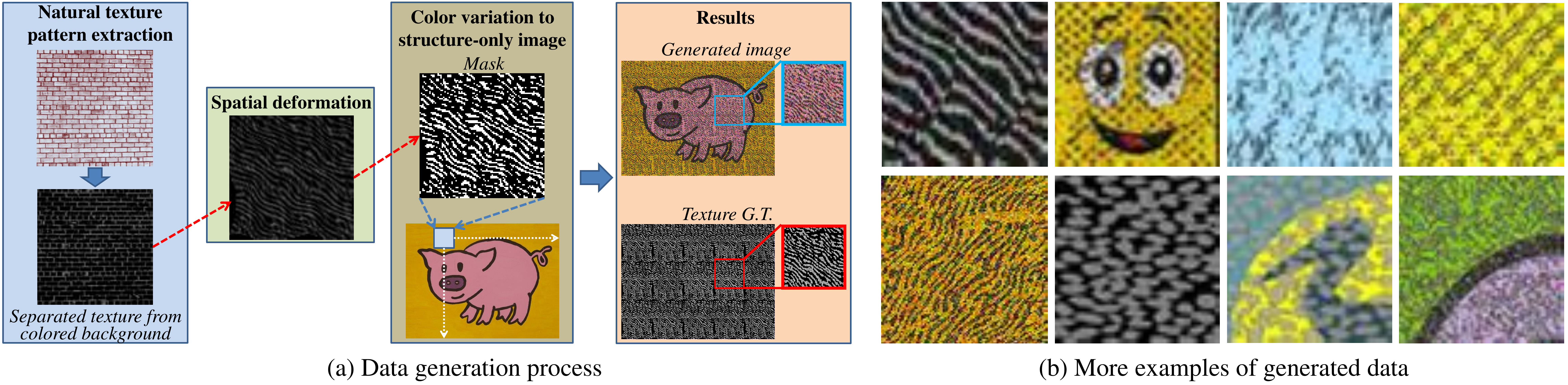}\\
\vspace{-12pt}
\caption{Illustration of dataset generation. We blend-in natural texture patterns to structure-only images, adding spatial and color variation to increase texture diversity.}
\label{fig:fig3}
\vspace{-10pt}
\end{figure*}

We observe that cartoon images have only structural edges filled with pure color, and can be safely considered ``structure-only images". Specifically, we select 174 cartoon images from the Internet and 233 different types of natural texture-only images from public datasets \cite{55cimpoi14describing,56dana1999reflectance}. The data generation process is illustrated in Fig.~\ref{fig:fig3}(a). Note that texture images in these datasets show textures only and all have simple backgrounds, so that separating them from the colored background is simple and efficient even using Relative Total Variation (RTV) \cite{23xu2012structure}. The texture layer separated by RTV is normalized to $[0,1]$.

Texture itself can be irregular, and textures in the wild may be distorted because of geometric projection. This arises because textures can appear on planar surfaces that are not orthogonal to the viewing direction, as well as being projected onto object with complex 3D surfaces. Therefore, we apply both spatial and color variation to the regular textures during dataset generation. As shown in Fig. \ref{fig:fig3}(a), we blend-in the texture to the structure-only image. In detail, we rescale all the texture images to $100 \times 100$ and extract texture patterns with RTV. We model spatial variation, capturing projected texture at patch level by performing geometric transforms including rotation, scaling, shearing, and linear and non-linear distortion. We randomly select the geometric transform and parameters for the operation\footnote{The detailed process can be found in the supplementary material, and we will provide the data generation code upon publication.}. Based on the deformed result, we generate a binary mask ${\bf{M}}$.

As for color variation, given the structure-only image ${\bf{S}}$, the value of pixel $i$ in the ${j^{th}}$ channel of the generated image ${\bf{I}}_i^{(j)}$ is determined by both ${\bf{S}}$ and the mask ${\bf{M}}$. If ${{\bf{M}}_i} = 1$, ${\bf{I}}_i^{(j)} = rand[\kappa  \cdot (1 - {\bf{S}}_i^{(j)}),1 - {\bf{S}}_i^{(j)}]$, where $\kappa$ is used to control the range of random generation and empirically set as 0.75. Otherwise, ${\bf{I}}_i^{(j)} = {\bf{S}}_i^{(j)}$. We repeat this by sliding the mask over the whole image without overlapping. The ground-truth texture confidence is calculated by averaging the values of the three channels of the texture layer:

\begin{equation}
{\bf{T}}_i^* = \delta (\frac{1}{3}\sum\limits_{j = 1}^3 {\left| {{\bf{I}}_i^{(j)} - {\bf{S}}_i^{(j)}} \right|} ),
\end{equation}
\noindent
where $\delta ( \cdot )$ is the sigmoid function to scale the value to $[0,1]$. We use this color variation to generate significant contrast between the textures and the background. Using this method, it is unlikely that two images have the same textures even when the textures come from the same original pattern. Fig.~\ref{fig:fig3}(b) shows eight generated image patches. 

Finally, we generate 30,000 images in total (a handful of low-quality images have been manually removed). For ground-truth, besides the purely-clean structure-only images, we also provide binary structure maps and texture confidence maps of all the generated images\footnote{More examples in the dataset are provided in the supplementary material, and the dataset will be available to the public upon publication.}.  

\begin{figure*}[tb]
\centering
\includegraphics[width=\linewidth]{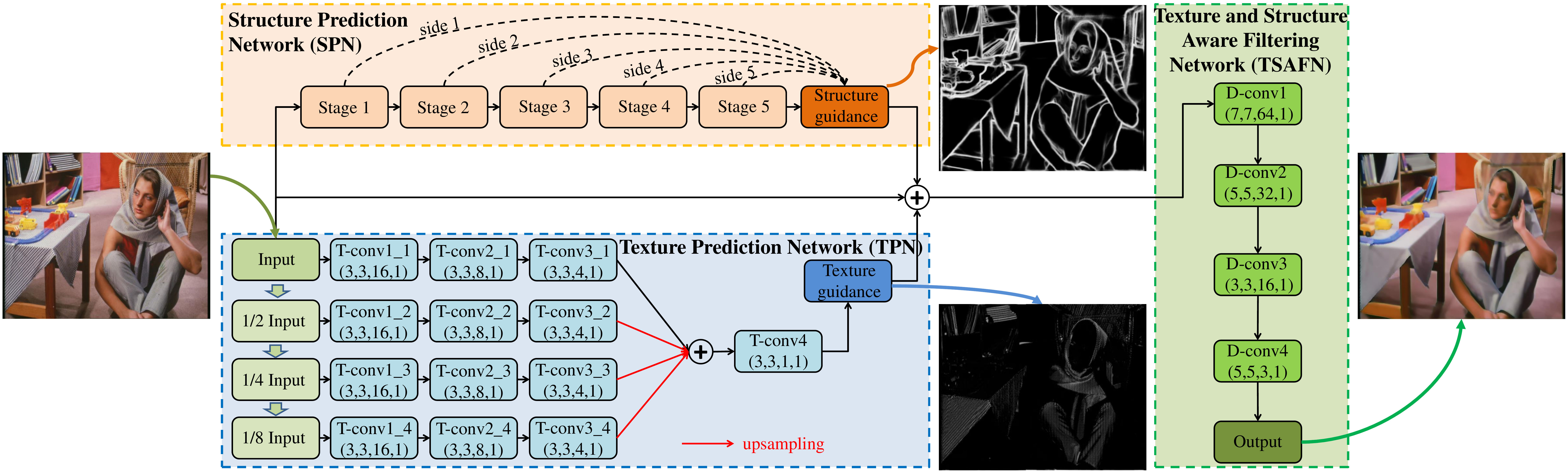}\\
\vspace{-9pt}
\caption{Our proposed network architecture. The outputs of the texture prediction network (TPN) and structure prediction network (SPN) are concatenated with the original input, and then fed to the texture and structure aware filtering network (TSAFN) to produce the final smoothing result. ($k$,$k$,$c$,$s$) for a convolutional layer means the kernel is $k \times k$ in size with $c$ feature maps, and the stride is $s$.}
\label{fig:network}
\vspace{-15pt}
\end{figure*}

\vspace{-7pt}
\subsection{Texture prediction network}
\vspace{-3pt}
\paragraph{\textbf{Network design}} We propose the texture prediction network (TPN), which is trained on our generated dataset. Considering that textures have various colors, scales, and shapes, we employ a multi-scale learning strategy.
Specifically, we apply 1/2, 1/4, and 1/8 down-sampling to the input respectively. For each image, we use 3 convolutional layers for feature extraction, with the same size $3 \times 3$ kernel and different number of feature maps. Then, all the feature maps are resized to the original input size and concatenated to form a 16-channel feature map. They are further convolved with a $3 \times 3$ layer to yield the final 1-channel result. Note that each convolutional layer is followed by ReLU except for the output layer, which is followed by a sigmoid activation function to scale the values to $[0,1]$. The architecture of TPN is shown in Fig.~\ref{fig:network}. Consequently, given the input image ${\bf{I}}$, the predicted texture guidance ${\bf{\tilde T}}$ is obtained by:

\begin{equation}
{\bf{\tilde T}} = g\left({\bf{I}},\frac{1}{2}{\bf{I}},\frac{1}{4}{\bf{I}},\frac{1}{8}{\bf{I}}\right).
\end{equation}

\paragraph{\textbf{Network training}} The network is trained by minimizing the mean squared error (MSE) between the predicted texture guidance map and the ground-truth:

\begin{equation}
{\ell _T}({\bf{\theta }}) = \frac{1}{N}\sum\limits_i {\left\| {{{{\bf{\tilde T}}}_i} - {\bf{T}}_i^*} \right\|_2^2},
\end{equation}

\noindent
where $N$ is the number of pixels in the image, $*$ denotes the ground-truth, and $\bf{\theta }$ represents parameters. More training details can be found in the experiment section.

\begin{figure}[tb]
\centering
\includegraphics[width=\linewidth]{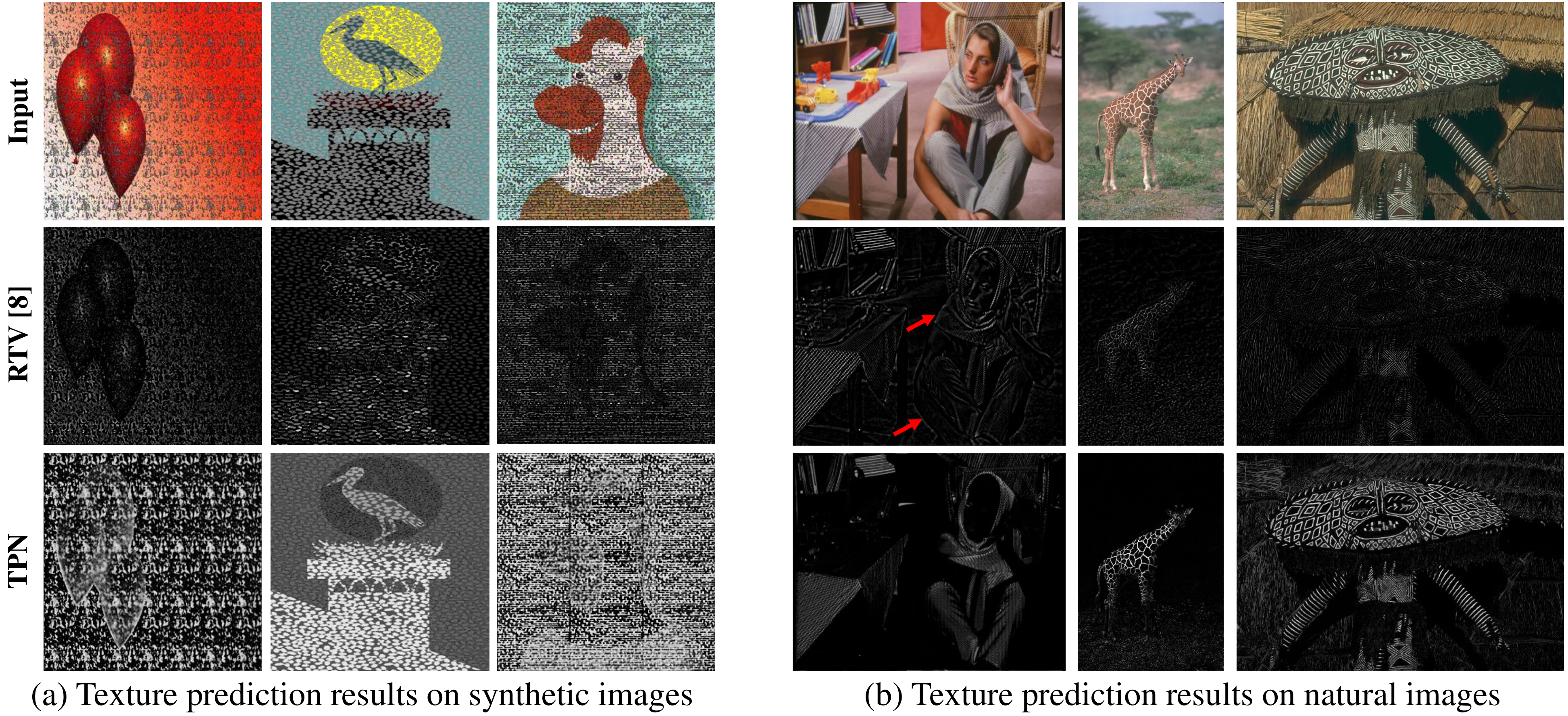}\\
\vspace{-10pt}
\caption{Texture prediction results. First row: input (including both generated and natural images). Second row: texture extraction results by RTV \cite{23xu2012structure} (we compare it because we use it to extract textures from texture-only images). Third row: texture prediction results by our proposed TPN. The network is able to find textures in both generated and natural images effectively, and indicate the magnitude of textures by assigning pixel-level confidence. RTV performs worse in extracting textures because just like other hand-crafted filters, it also assumes structures have large gradients and has poor discrimination of strong textures, especially in more complicated scenes.}
\label{fig:texturePrediction}
\vspace{-10pt}
\end{figure}

\vspace{-7pt}
\paragraph{\textbf{Texture prediction results}}
We present the texture prediction results on our generated images in Fig.~\ref{fig:texturePrediction}(a) and natural images in Fig.~\ref{fig:texturePrediction}(b). The network is able to find textures in both the generated and natural images effectively, and indicate the magnitude of textures by assigning pixel-level confidence (the third row). For comparison, we also list the texture extraction results from these examples by RTV \cite{23xu2012structure} in the second row. RTV performs worse on the more complex scenes, and some structures are unexpectedly visible in the texture layer (red arrows). This is because just like other hand-crafted filters, RTV also assumes structures have large gradients and hence has poor discrimination of strong textures.

\section{Texture and Structure Aware Filtering Network}
\vspace{-7pt}
As shown in Fig.~\ref{fig:network}, our deep filtering network consists of three parts:
\begin{enumerate}
 \item Texture prediction network \textbf{TPN}, that constructs texture guidance to indicate texture regions and magnitude (texture confidence).
  \item Structure prediction network \textbf{SPN}, that constructs structure guidance to indicate meaningful structures (structure confidence).
  \item Texture and structure aware filtering network \textbf{TSAFN}, that concatenates the two guidance images with the original input and generates the smoothing output.
\end{enumerate}
\noindent
Since TPN has been discussed in the previous section, we give more details about SPN and TSAFN in the following.

\vspace{-7pt}
\subsection{Structure prediction network}
Structure information is an essential cue for image smoothing, that tells the filter which boundaries should be preserved. The ideal structure guidance would give high confidence to meaningful structures, regardless of gradient intensity. We utilize a recently-proposed holistically-nested edge detection (HED) \cite{54xie2015holistically} as the structure prediction network (SPN):

\begin{equation}
{\bf{\tilde E}} = f({\bf{I}}) = {\rm{fuse(}}{{{\bf{\tilde E}}}^{(1)}}{\rm{,}}...{\rm{,}}{{{\bf{\tilde E}}}^{(5)}}{\rm{)}},
\end{equation}
\noindent
where ${{{\bf{\tilde E}}}^{(m)}}$ is the side output from the ${m^{th}}$ stage (each stage contains several convolutional and pooling layers). The final loss is denoted as ${\ell _E}({\bf{\theta }})$. Please refer to the original paper \cite{54xie2015holistically} for more details.

\vspace{-7pt}
\subsection{Texture and structure aware filtering network}
\label{sect:network}
\vspace{-5pt}
\paragraph{\textbf{Network design}}
Once the structure and texture guidance are generated, the texture and structure aware filtering network (TSAFN) concatenates them with the input to form a 5-channel tensor. TSAFN consists of 4 layers. We set a relatively large kernel ($7 \times 7$) in the first layer to take more original information into account. The kernel size decreases in the following two layers ($5 \times 5$, $3 \times 3$ respectively). In the last layer, the kernel size is increased to $5 \times 5$ again. The first three layers are followed by ReLU, while the last layer has no activation function (transforming the tensor into the 3-channel output). Empirically, we remove all the pooling layers, the same as \cite{46xu2015deep,48li2016deep,49Fan_2017_ICCV,52Chen_2017_ICCV}. We set the filtering network without any guidance as the baseline. The whole process can be denoted as:

\begin{equation}
{\bf{\tilde I}} = h({\bf{I}},{\bf{\tilde E}},{\bf{\tilde T}}).
\end{equation}

\paragraph{\textbf{Network training}}
The network is trained by minimizing:
\begin{equation}
{\ell _D}({\bf{\theta }}) = \frac{1}{N}\sum\limits_i {(\left\| {{{{\bf{\tilde I}}}_i} - {\bf{I}}_i^*} \right\|_2^2)}.
\end{equation}
\noindent
More details can be found in the experiment section.

\vspace{-7pt}
\section{Experiments and Analysis}
In this section, we demonstrate the effectiveness of our proposed deep image smoothing network through.
\vspace{-5pt}
\paragraph{\textbf{Environment setup}}
We construct the networks in Tensorflow \cite{62abadi2016tensorflow}, and train and test all the data on a single NVIDIA Titan X graphics card. 
\vspace{-5pt}
\paragraph{\textbf{Dataset}}
Because there is no existing texture removal dataset, we perform training using our generated images. More specifically, we select 19,505 images (65\%) from the dataset for training, 2,998 (10\%) for validation, and 7,497 (25\%) for testing (all test images are resized to $512 \times 512$). There is no overlapping of the structure-only images between training, validation and testing samples. 
 \vspace{-5pt}
\paragraph{\textbf{Training}}
We first train the three networks separately. 300,000 patches with the size $64 \times 64$ are randomly and sparsely collected from training images. We use gradient descent with a learning rate of 0.0001, and momentum of 0.9. Finally, we perform fine-tuning by jointly training the whole network with a smaller learning rate of 0.00001, and the same momentum 0.9. The fine-tuning loss is
\begin{equation}
\ell ({\bf{\theta }}) = \gamma  \cdot {\ell _D}({\bf{\theta }}) + \lambda  \cdot ({\ell _T}({\bf{\theta }}) + {\ell _E}({\bf{\theta }})),
\end{equation}

\noindent
where we empirically set $\gamma  = 0.6$, and $\lambda  = 0.2$.
\begin{figure*}[!t]
\centering
\subfigure{\includegraphics[width=\linewidth]{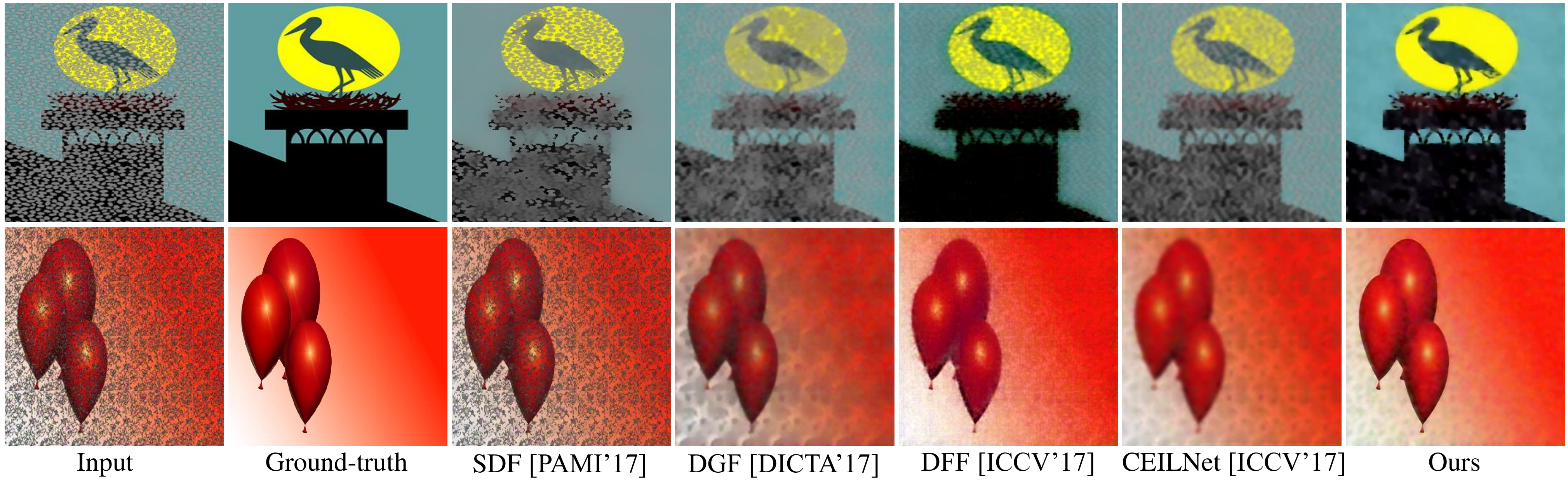}}
\\
\vspace{-12pt}
\caption{
Smoothing results on generated images. Our filter can smooth out various types of textures while preserving structures more effectively than other approaches.
}
\label{fig:synTesting}
\vspace{-14pt}
\end{figure*}

\vspace{-7pt}
\subsection{Existing methods to compare}
\vspace{-2pt}
\paragraph{\textbf{Traditional hand-crafted methods}}
We compare our filter with 2 classical filters: Total Variation (TV) \cite{22rudin1992nonlinear}, bilateral filter (BLF) \cite{1tomasi1998bilateral}, and 9 state-of-the-art filters: L0 \cite{4xu2011image}, Relative Total Variation (RTV) \cite{23xu2012structure}, guided filter (GF) \cite{2he2013guided}, Structure Gradient and Texture Decorrelation (SGTD) \cite{9liu2013sgtd}, rolling guidance filter (RGF) \cite{11zhang2014rolling}, fast L0 \cite{26nguyen2015fast}, segment graph filter (SGF) \cite{3zhang2015segment}, static and dynamic filter (SDF) \cite{13ham2017robust}, double-guided filter (DGF) \cite{53Lu_2017_DICTA}. Note that, BLF, GF, RGF, SGF, DGF are kernel-based, while TV, L0, RTV, SGTD, fast L0, SDF are separation-based. We use the default parameters defined in the open-source code for each method.
\vspace{-7pt}
\paragraph{\textbf{Deep learning based methods}}
We select 5 state-of-the-art deep filtering models: deep edge-aware filter (DEAF) \cite{46xu2015deep}, deep joint filter (DJF) \cite{48li2016deep}, deep recursive filter (DRF) \cite{47liu2016learning}, deep fast filter (DFF) \cite{52Chen_2017_ICCV}, and cascaded edge and image
learning network (CEILNet) \cite{49Fan_2017_ICCV}
. \textbf{We retrain all the models with our dataset}.

\vspace{-10pt}
\subsection{Results}
\vspace{-3pt}
\paragraph{\textbf{Quantitative results on generated images}}
We first compare the average MSE, PSNR, SSIM \cite{44wang2004image}, and processing time (in seconds) of 11 hand-crafted filters on our testing data in Table~\ref{tab:differentfilter}. Our method achieves the smallest MSE (closest to ground-truth), largest PSNR and SSIM (removing textures and preserving main structures most effectively), and lowest running time, indicating its superiority in both effectiveness and efficiency. Note that although the double-guided filter (DGF) \cite{53Lu_2017_DICTA} achieves better quantitative results than other hand-crafted approaches, it runs extremely slowly (more than \textit{50 times} slower than ours). This may be due to the complex process of generating two guidances, and the inefficiency of the kernel operation. We also compare the quantitative results on different deep models trained and tested on our dataset in Table~\ref{tab:tab2}. Our model achieves the best MSE, PSNR and SSIM, with comparable efficiency to the other methods. We additionally select 4 state-of-the-art methods (SDF \cite{13ham2017robust}, DGF \cite{53Lu_2017_DICTA}, DFF \cite{52Chen_2017_ICCV}, and CEILNet \cite{49Fan_2017_ICCV}) for visual comparison in Fig.~\ref{fig:synTesting}. The textures in the first example have relatively large scale. SDF, DGF, and CEILNet attempt to remove these textures but the structures are blurred severely as a penalty. In the second example, the textures are densely distributed and have relatively large contrast. SDF performs badly in this example due to the poor texture discrimination. DGF and CEILNet can suppress these textures, but the structures are blurred. Although DFF is able to smooth out almost all the textures, the final results show unexpected artifacts and color shift, and look less similar to the ground-truth than ours. Only our filter performs well in both examples.

\begin{figure*}[!t]
\centering
\subfigure{\includegraphics[width=\linewidth]{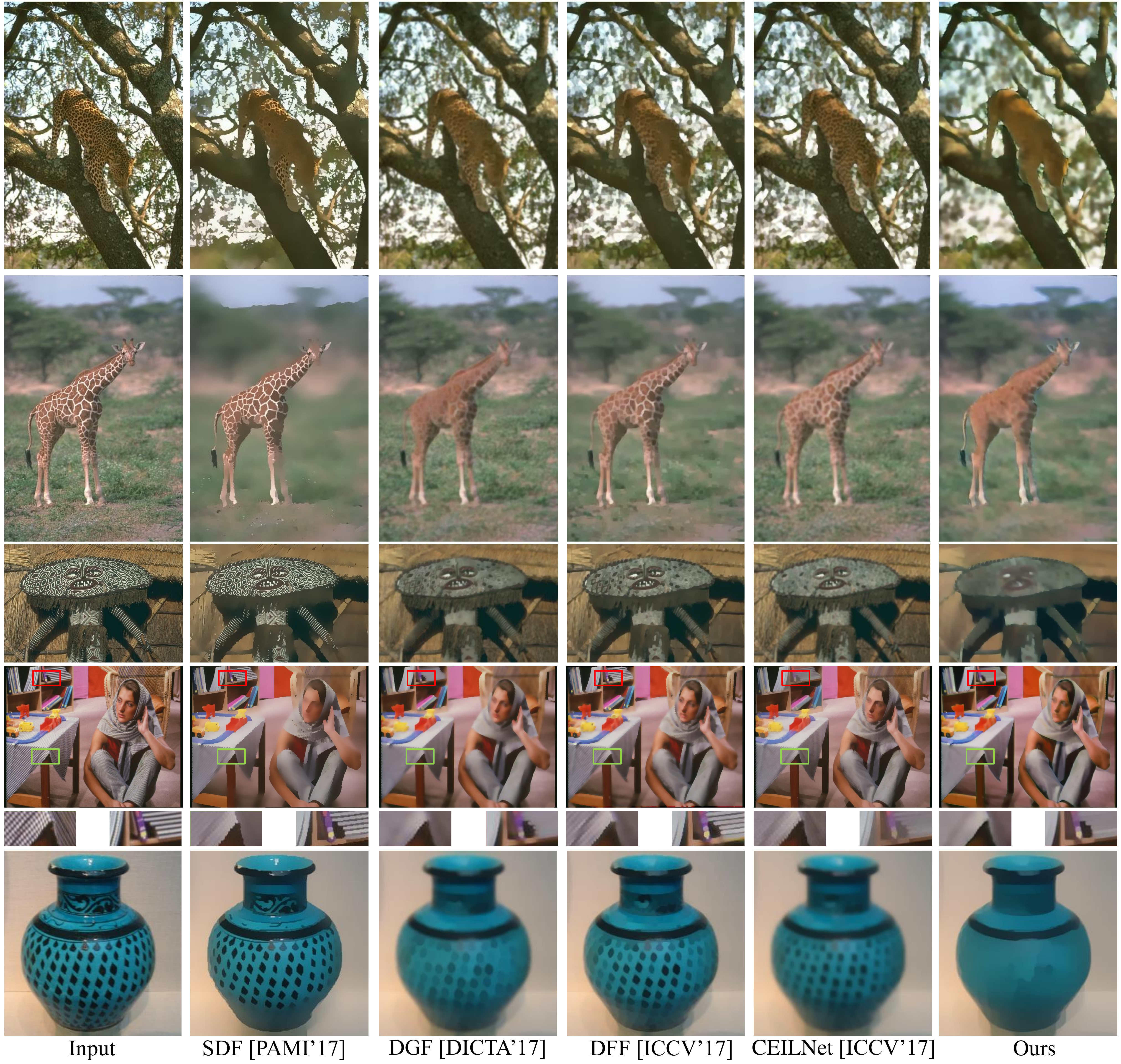}}\\
\vspace{-12pt}
\caption{
Smoothing results on natural images. The first example shows the ability of weak structure preservation and enhancement in textured scenes. The next four examples present various texture types with different shapes, contrast, and distortion. Our filter performs consistently better than state-of-the-art methods in all the examples, demonstrating its superiority in image smoothing and good generality in processing natural images.
}
\label{fig:natural}
\vspace{-15pt}
\end{figure*}

\begin{table}[ht]
\small
\caption{Quantitative evaluation of different hand-crafted filters tested on our dataset}
\label{tab:differentfilter}
\begin{center}
\begin{tabular}{|c|c|c|c|c|c|c|c|c|c|} 
\hline
\rule[-1ex]{0pt}{3.5ex}   & MSE &PSNR &SSIM &Time & & MSE &PSNR &SSIM &Time    \\
\hline\hline
\rule[-1ex]{0pt}{3.5ex}  TV \cite{22rudin1992nonlinear} &0.2791  &11.33 &0.6817 & 2.44   &\rule[-1ex]{0pt}{3.5ex}  RGF \cite{11zhang2014rolling} &0.2094  &15.73 &0.7173 & 0.87\\
\hline
\rule[-1ex]{0pt}{3.5ex}  BLF \cite{1tomasi1998bilateral} &0.3131  &10.89 &0.6109 & 4.31  &\rule[-1ex]{0pt}{3.5ex}  Fast L0 \cite{26nguyen2015fast} &0.2068  &15.50 & 0.7359 &  1.36 \\
\hline
\rule[-1ex]{0pt}{3.5ex}  L0 \cite{4xu2011image} &0.2271  &14.62 & 0.7133 &  0.94 &\rule[-1ex]{0pt}{3.5ex}  SGF \cite{3zhang2015segment} &0.2446  &13.92 & 0.7002  & 2.26\\
\hline
\rule[-1ex]{0pt}{3.5ex}  RTV \cite{23xu2012structure} &0.2388  &14.07 & 0.7239  & 1.23 &\rule[-1ex]{0pt}{3.5ex}  SDF \cite{12ham2015robust} &0.1665  &16.82 &0.7633   & 3.71\\
\hline
\rule[-1ex]{0pt}{3.5ex}  GF \cite{2he2013guided} &0.2557  &12.22 &0.6948   & 0.83  &\rule[-1ex]{0pt}{3.5ex}  DGF \cite{53Lu_2017_DICTA} &0.1247  &17.89 &0.7552   & 8.66\\
\hline
\rule[-1ex]{0pt}{3.5ex}  SGTD \cite{9liu2013sgtd} &0.1951  &16.14 &0.7538 & 1.59   &\rule[-1ex]{0pt}{3.5ex}  Ours &\textbf{0.0051}  &\textbf{25.07} &\textbf{0.9152}   & \textbf{0.16}\\
\hline
\end{tabular}
\end{center}
\vspace{-10pt}
\end{table}

\vspace{-5pt}
\paragraph{\textbf{Qualitative comparison on real images in the wild}}

\begin{table}[ht]
\small
\caption{Quantitative evaluation of deep models trained and tested on our dataset}
\label{tab:tab2}
\begin{center}
\vspace{-5pt}
\begin{tabular}{|c|c|c|c|c|c|c|c|c|c|} 
\hline
\rule[-1ex]{0pt}{3.5ex}   & MSE &PSNR &SSIM &Time  & & MSE &PSNR &SSIM &Time   \\
\hline\hline
\rule[-1ex]{0pt}{3.5ex}  DEAF \cite{46xu2015deep} &0.0297  &20.62 &0.8071 & 0.35 & \rule[-1ex]{0pt}{3.5ex}  DFF \cite{52Chen_2017_ICCV} &0.0172  &22.21 & 0.8675  & \textbf{0.07}\\
\hline
\rule[-1ex]{0pt}{3.5ex}  DJF \cite{48li2016deep} &0.0352  &19.01 &0.7884 & 0.28   &\rule[-1ex]{0pt}{3.5ex}  CEILNet \cite{49Fan_2017_ICCV} &0.0156  &22.65 & 0.8712  & 0.13\\
\hline
\rule[-1ex]{0pt}{3.5ex}  DRF \cite{47liu2016learning} &0.0285  &21.14 & 0.8263 &  0.12 &\rule[-1ex]{0pt}{3.5ex}  Ours &\textbf{0.0051}  &\textbf{25.07} &\textbf{0.9152}   & 0.16\\
\hline
\end{tabular}
\end{center}
\vspace{-18pt}
\end{table}

\begin{table}[ht]
\small
\caption{Ablation study of image smoothing effects with no guidance, only structure guidance, only texture guidance, and double guidance (trained separately and fine-tuned)}
\label{tab:tab1}
\begin{center}
\begin{tabular}{|c|c|c|c|} 
\hline
\rule[-1ex]{0pt}{3.5ex}   & MSE &PSNR &SSIM     \\
\hline\hline
\rule[-1ex]{0pt}{3.5ex}  No guidance (Baseline) &0.0316  &20.32 &0.7934   \\
\hline
\rule[-1ex]{0pt}{3.5ex}  Only structure guidance &0.0215  &21.71 &0.8671     \\
\hline
\rule[-1ex]{0pt}{3.5ex}  Only texture guidance &0.0118  &23.23 & 0.8201   \\
\hline
\rule[-1ex]{0pt}{3.5ex}  Double guidance (trained separately) &0.0059  &24.78 &0.9078    \\
\hline
\rule[-1ex]{0pt}{3.5ex}  Double guidance (fine-tuned) &\textbf{0.0051}  &\textbf{25.07} &\textbf{0.9152}    \\
\hline
\end{tabular}
\end{center}
\vspace{-10pt}
\end{table}

We visually compare smoothing results of 5 challenging natural images with SDF \cite{13ham2017robust}, DGF \cite{53Lu_2017_DICTA}, DFF \cite{52Chen_2017_ICCV}, and CEILNet \cite{49Fan_2017_ICCV} in Fig.~\ref{fig:natural}. In the first example, the leopard is covered with black texture, and it has relatively low contrast to the background (weak structure). Only our filter smooths out all the textures while effectively preserving and enhancing the structure. The next four examples present various texture types with different shapes, contrast, and distortion. Our filter performs consistently well in both preserving structures and removing textures. We analyze the last challenging vase example in more detail. The vase is covered with strong dotted textures, densely distributed on the surface. SDF fails to remove these textures since they are regarded as structures with large gradients. DGF smooths out the black dots more effectively but the entire image looks blurry. This is because just as \cite{53Lu_2017_DICTA} points out, a larger kernel size and more iterations are required to remove more textures, resulting in the blurred structure as a penalty. Also, the naive combination of structure and texture kernels makes the filter not robust to various types of textures, in which case the structure may not always be retained well even with the proper structure guidance. The two deep filters do not demonstrate much improvement over the hand-crafted approaches because ``texture-awareness" is not specially emphasized in their network design, necessitating a trade-off between structure preservation and texture removal. Only our filter removes all the textures without blurring the main structure.

\begin{figure*}[!t]
\centering
\subfigure{\includegraphics[width=\linewidth]{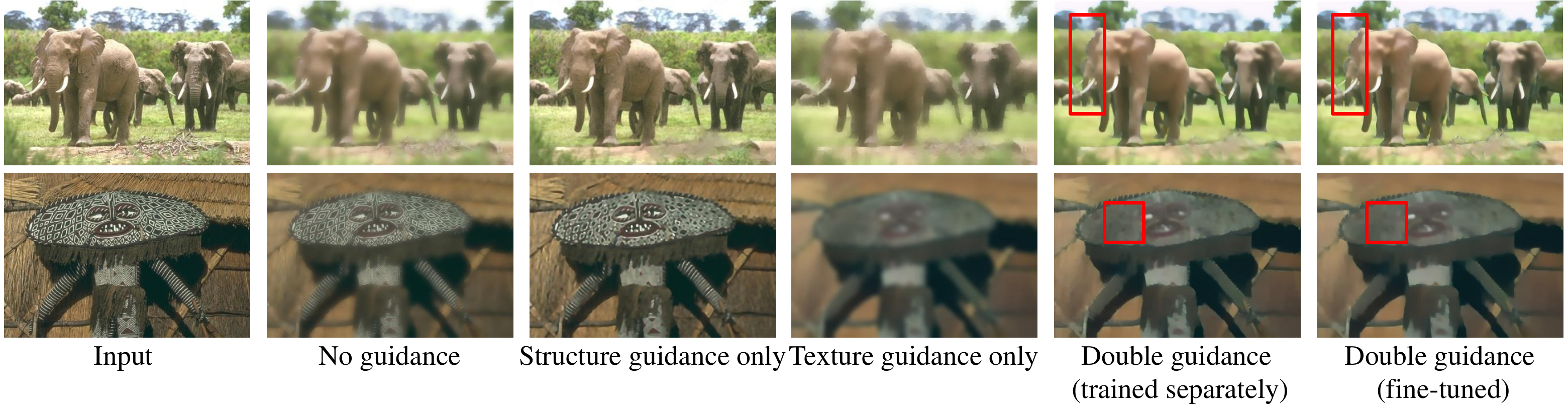}}\\
\vspace{-12pt}
\caption{
Image smoothing results with no guidance, single guidance, double guidance (trained separately, and fine-tuned). With only structure guidance, the main structures are retained as well as the textures. With only texture guidance, all the textures are smoothed out but the structures are severely blurred. The result with double guidance performs well in both structure preservation and texture removal. Fine-tuning the whole network can further improve the performance.
}
\label{fig:ablation}
\vspace{-13pt}
\end{figure*}

\vspace{-5pt}
\paragraph{\textbf{Ablation study of each guidance}} To investigate the effect of guidance, we train the filtering network with no guidance, only structure guidance, only texture guidance, and double guidance respectively. We list the average MSE, PSNR, and SSIM of the testing results compared with ground-truth in Table~\ref{tab:tab1}, demonstrating that the results with double guidance have smaller MSE, larger PSNR, and larger SSIM. Also, the fine-tuning process improves the filtering network. Further, we show two natural images in Fig.~\ref{fig:ablation}. Compared with the baseline without guidance, the result only with structure guidance retains more structure, as well as the texture (this is mainly because HED may also be negatively affected by strong textures, resulting in a larger MSE loss when training the network). In contrast, the structures are severely blurred with only texture guidance, even though most textures are removed. Combining both structure and texture guidance produces a better result. Fine-tuning further improves the result (in the red rectangle of the first example, the structures are sharper; in the second example, the textures within the red region are further suppressed). All the observations are consistent with the quantitative evaluation in Table~\ref{tab:tab1}.

\vspace{-12pt}
\section{Conclusion}
\vspace{-9pt}
In this paper, we propose an end-to-end texture and structure aware filtering network that is able to smooth images with both ``texture-awareness" and ``structure-awareness". The ``texture-awareness" benefits from the newly-proposed texture prediction network. To facilitate training, we blend-in natural textures onto structure-only cartoon images with spatial and color variations. The ``structure-awareness" is realized by semantic edge detection. Experiments show that the texture network can detect textures effectively. And our filtering network outperforms other kernel-based, separation-based, and learning-based filters on both generated images and natural images. The network structure is intuitive and easy to implement, and achieves excellent smoothing ability with comparable efficiency to state-of-the-art methods.

\clearpage

\bibliographystyle{splncs}
\bibliography{egbib}

\clearpage
\begin{center}
\textbf{Supplementary Material}
\end{center}

\section{Overview}

The purpose of this supplementary material is to provide more analysis of our method and experimental results. Specifically, we first give more details about adding spatial variation to training data. Then, we provide the details about how to train the structure prediction network. After that, we give more examples of image smoothing results, as well as qualitative comparison with other methods. We also present a challenging case and analyze potential reasons behind it. Finally, we apply our method to, and show results for, three typical applications of image smoothing.

\section{Details of Adding Spatial Variations to Training Data}
Textures can appear in spatially varying forms in natural images as the texture is formed over objects and projected by the imaging process. We add this property when generating our training data. We mainly use four types of geometric transformation in the generating process: scaling, shearing, rotation, and free-form distortion. According to the geometric theory of computer vision \cite{65artley2003multiple}, combining the first three operations can be used to yield a \textit{weak perspective projection}, which is consistent with the formation of natural images by cameras (assuming the camera is not too close to the scene). Free-form distortion is used to represent projection onto more arbitrary surfaces such as the body of a giraffe, or texture printed on a sheet of waving material.

\begin{figure*}[!t]
\centering
\subfigure{\includegraphics[width=\linewidth]{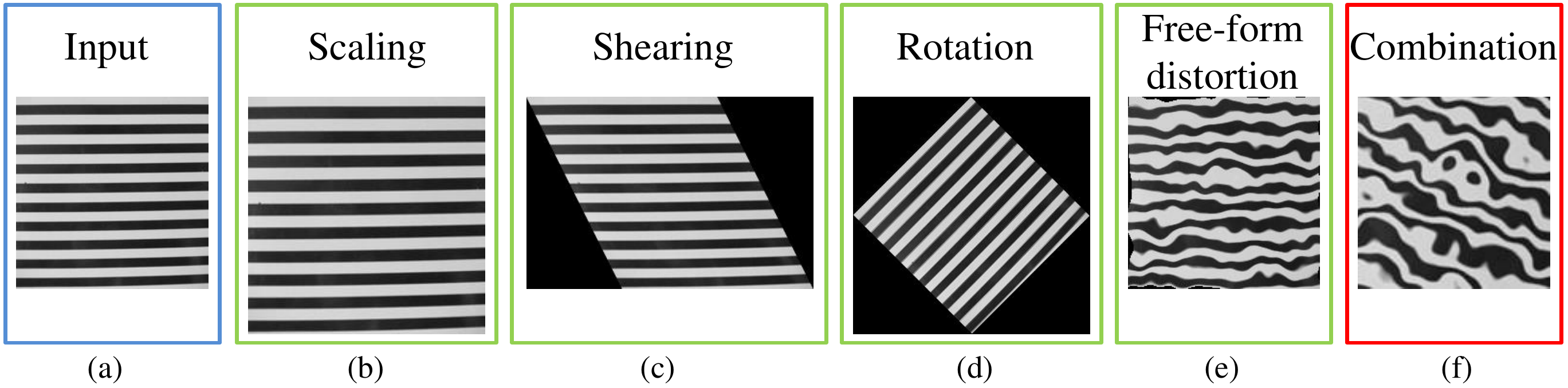}}\\
\caption{
Illustration of spatial variations.
}
\label{fig:spatial}
\end{figure*}

Suppose the original coordinate is $(x,y)$, and the transformed one is $(x',y')$, we give formulas in the following. We also give a stripe texture example for illustration in Fig.~\ref{fig:spatial}.

\paragraph{\textbf{Scaling:}}resizes an image in $x$ or/and $y$ directions:
\begin{equation}
\left[ {\begin{array}{*{20}{c}}
{x'}\\
{y'}
\end{array}} \right] = \left[ {\begin{array}{*{20}{c}}
{{s_1}}&0\\
0&{{s_2}}
\end{array}} \right]\left[ {\begin{array}{*{20}{c}}
x\\
y
\end{array}} \right].
\end{equation}

\noindent
In our implementation, both ${s_1}$ and ${s_2}$ are randomly generated and fall into the range of $(1,3]$, as shown in Fig.~\ref{fig:spatial}(b).

\paragraph{\textbf{Shearing:}}stretches the image in $x$ or/and $y$ directions:
\begin{itemize}
\item $x$ direction:
\begin{equation}
\left[ {\begin{array}{*{20}{c}}
{x'}\\
{y'}
\end{array}} \right] = \left[ {\begin{array}{*{20}{c}}
1&{{k_1}}\\
0&1
\end{array}} \right]\left[ {\begin{array}{*{20}{c}}
x\\
y
\end{array}} \right].
\end{equation}
\item $y$ direction:
\begin{equation}
\left[ {\begin{array}{*{20}{c}}
{x'}\\
{y'}
\end{array}} \right] = \left[ {\begin{array}{*{20}{c}}
1&0\\
{{k_2}}&1
\end{array}} \right]\left[ {\begin{array}{*{20}{c}}
x\\
y
\end{array}} \right].
\end{equation}
\end{itemize}

\noindent
In our implementation, ${k_1}$ and ${k_2}$ are randomly generated and fall into the range of $[0,1]$, as shown in Fig.~\ref{fig:spatial}(c).

\paragraph{\textbf{Rotation:}}rotates an image in the image plane by an angle $\theta $:

\begin{equation}
\left[ {\begin{array}{*{20}{c}}
{x'}\\
{y'}
\end{array}} \right] = \left[ {\begin{array}{*{20}{c}}
{\cos \theta }&{\sin \theta }\\
{ - \sin \theta }&{\cos \theta }
\end{array}} \right]\left[ {\begin{array}{*{20}{c}}
x\\
y
\end{array}} \right].
\end{equation}

\noindent
In our implementation, $\theta $ is randomly generated in the range of $[ - \pi ,\pi ]$, as shown in Fig.~\ref{fig:spatial}(d).

\paragraph{\textbf{Free-form distortion:}}we add a free-form distortion in spatial coordinates to make the textures look more like in natural scenes. In our method, we randomly switch pixel values within a $f\times f$ kernel $\Omega$:

\begin{equation}
\left[ {\begin{array}{*{20}{c}}
{x'}\\
{y'}
\end{array}} \right] = rand\{ \Omega (x,y)\}.
\end{equation}

\noindent
In our implementation, $f$ is the kernel size and randomly selected from $\{ 3,5,7,9,11\}$, as shown in Fig.~\ref{fig:spatial}(e).

One possible combined result is shown in Fig.~\ref{fig:spatial}(f). Compared with the original input, it has more variation in spatial coordinates, which is closer to some of the more distorted textures that we see in the nature.

\section{Details of Training Structure Prediction Network}

\paragraph{\textbf{Network architecture}}  We use the HED \cite{54xie2015holistically} architecture to construct the structure prediction network (SPN). As shown in Fig.~\ref{fig:fig3}, SPF has 5 stages (each stage contains several convolutional and pooling layers) and is embedded in VGGNet (trims it by adding side output to the last convolutional layer at each stage, and replaces the fully connected layers with fully convolutional layers at the last stage). The side outputs are then fused to form the final output. The whole process can be expressed as the following function:

\begin{equation}
{\bf{\tilde E}} = f({\bf{I}}) = {\rm{fuse(}}{{{\bf{\tilde E}}}^{(1)}}{\rm{,}}...{\rm{,}}{{{\bf{\tilde E}}}^{(5)}}{\rm{)}},
\end{equation}
\noindent
where ${{{\bf{\tilde E}}}^{(m)}}$ is the side output from the ${m^{th}}$ stage (each stage contains several convolutional and pooling layers). Please find more details about the architecture in the original paper \cite{54xie2015holistically} and published code\footnote{https://github.com/ppwwyyxx/tensorpack/tree/master/examples/HED}.

\paragraph{\textbf{Network training}} 
During training, we randomly and sparsely collect 300,000 patches with size $64 \times 64$ from the 25,176 training images (as mentioned in the main body). Since we have binary edge maps, we follow the steps in \cite{54xie2015holistically} to re-train the network by considering both side output loss and fusion loss. The side loss of the ${m^{th}}$ stage is defined as

\begin{equation}
\ell _{side}^{(m)}({\bf{\theta }}) =  - \beta \sum\limits_{i \in {\bf{E}}_ + ^*} {\log ({\bf{\tilde E}}_i^{(m)})}  - (1 - \beta )\sum\limits_{i \in {\bf{E}}_ - ^*} {\log (1 - {\bf{\tilde E}}_i^{(m)})},
\end{equation}
where ${{\bf{E}}_ + ^*}$ and ${\bf{E}}_ - ^*$ denote the edge (1) and non-edge (0) ground-truth labels respectively, 
$\beta  = {{\left| {{\bf{E}}_ + ^*} \right|} \mathord{\left/
 {\vphantom {{\left| {{\bf{E}}_ + ^*} \right|} {\left| {{{\bf{E}}^*}} \right|}}} \right.
 \kern-\nulldelimiterspace} {\left| {{{\bf{E}}^*}} \right|}}$ represents the proportion of edge labels, and ${\bf{\theta }}$ is the set of parameters. 
The total side output loss ${\ell _{side}}({\bf{\theta }})$ is the sum of five stages:

\begin{equation}
{\ell _{side}}({\bf{\theta }}) = \sum\limits_{m = 1}^5 {\ell _{side}^{(m)}({\bf{\theta }})}.
\end{equation}

\noindent
The fusion loss ${\ell _{fuse}}({\bf{\theta }})$ is calculated by the cross entropy loss between the fused image and ground-truth:

\begin{equation}
{\ell _{fuse}}({\bf{\theta }}) =  - \beta \sum\limits_{i \in {\bf{E}}_ + ^*} {\log ({{{\bf{\tilde E}}}_i})}  - (1 - \beta )\sum\limits_{i \in {\bf{E}}_ - ^*} {\log (1 - {{{\bf{\tilde E}}}_i})}.
\end{equation}

\noindent
The total loss is the combination of side output loss and fusion loss:

\begin{equation}
{\ell _E}({\bf{\theta }}) = {\ell _{side}}({\bf{\theta }}) + {\ell _{fuse}}({\bf{\theta }}).
\end{equation}

\noindent
We replace the Adam optimizer with the gradient descent algorithm (learning rate 0.0001, and momentum 0.9).    

\section{Additional Experiments on Texture Extraction}
\paragraph{\textbf{Texture extraction results on our test data}} As shown in Fig.~\ref{fig:fig3}, when generating the data, we also provide ground-truth for texture prediction. Thus, we can investigate the texture extraction abilities of different methods by comparing the extracted textures with \textit{texture ground-truth} (we actually aim to compare our single TPN with other methods). We present the texture extraction abilities of our method along with 6 typical texture separation algorithms that we select for comparison for comparison: Total Variation (TV) \cite{22rudin1992nonlinear}, L0 \cite{4xu2011image}, Relative Total Variation (RTV) \cite{23xu2012structure}, Structure Gradient and Texture Decorrelation (SGTD) \cite{9liu2013sgtd}, fast L0 \cite{26nguyen2015fast}, static and dynamic filter (SDF) \cite{13ham2017robust}, and normalize their texture layers as the final results. We report the average MSE of different methods tested on our 7,497 testing data in Table~\ref{tab:supTab1}. Our TPN achieves the smallest MSE among all the methods, showing its superiority in extracting textures.

\begin{table}[ht]
\small
\caption{Quantitative evaluation of texture extraction results tested on our dataset}
\label{tab:supTab1}
\begin{center}
\vspace{-5pt}
\begin{tabular}{|c|c|c|c|c|c|c|c|} 
\hline
\rule[-1ex]{0pt}{3.5ex}     Methods &TV \cite{22rudin1992nonlinear} & L0 \cite{4xu2011image} & RTV \cite{23xu2012structure} & SGTD \cite{9liu2013sgtd} & Fast L0 \cite{26nguyen2015fast} & SDF \cite{13ham2017robust} & Ours\\
\hline
\rule[-1ex]{0pt}{3.5ex}     \makecell{MSE \\ (texture \\extraction)} &0.2175 & 0.2246 & 0.1954 & 0.1315 & 0.2369 & 0.1738 & \textbf{0.0196}\\
\hline
\rule[-1ex]{0pt}{3.5ex}     \makecell{MSE \\ (image \\smoothing)} &0.2791 & 0.2271 & 0.2388 & 0.1951 & 0.2068 & 0.1665 & \textbf{0.0051}\\
\hline
\end{tabular}
\end{center}
\vspace{-20pt}
\end{table}

\begin{table}[ht]
\small
\caption{Quantitative evaluation of texture extraction results tested on 100 new images}
\label{tab:supTab2}
\begin{center}
\vspace{-5pt}
\begin{tabular}{|c|c|c|c|c|c|c|c|} 
\hline
\rule[-1ex]{0pt}{3.5ex}     Methods &TV \cite{22rudin1992nonlinear} & L0 \cite{4xu2011image} & RTV \cite{23xu2012structure} & SGTD \cite{9liu2013sgtd} & Fast L0 \cite{26nguyen2015fast} & SDF \cite{13ham2017robust} & Ours\\
\hline
\rule[-1ex]{0pt}{3.5ex}     \makecell{MSE \\ (texture \\extraction)} &0.2331 & 0.2494 & 0.2017 & 0.1608 & 0.2433 & 0.1795 & \textbf{0.0212}\\
\hline
\rule[-1ex]{0pt}{3.5ex}     \makecell{MSE \\ (image \\smoothing)} &0.2880 & 0.2539 & 0.2375 & 0.1974 & 0.2342 & 0.2190 & \textbf{0.0074}\\
\hline
\end{tabular}
\end{center}
\vspace{-5pt}
\end{table}

\begin{figure*}[!t]
\centering
\subfigure{\includegraphics[width=\linewidth]{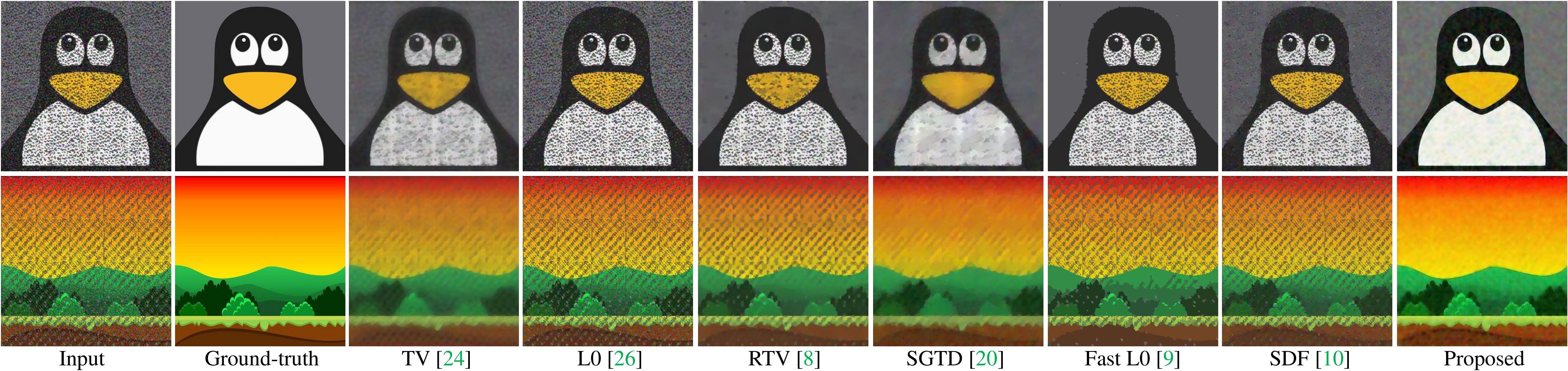}}\\
\caption{
Image smoothing results on new images.
}
\label{fig:newdata}
\end{figure*}

\paragraph{\textbf{Texture extraction results on a new dataset}} To further verify the generality of our proposed TPN to different types of textures, we make another small dataset specially for this testing. Specifically, we select 50 natural texture images from another public dataset\footnote{The dataset is from Signal and Image Processing Institute, University of Southern California. It is available at: http://sipi.usc.edu/database/database.php?volume=texturesimage=1top}, and 100 other structure-only cartoon images from the Internet. We blend-in these new textures to structure-only images in the same way as mentioned in the main body. In Table~\ref{tab:supTab2}, we report the average MSE tested on the 100 new images. Unsurprisingly, our TPN achieves the smallest MSE again, indicating its adaptation to different types of textures. This result also helps explain why our TPN and filtering networks generalize well to natural image processing. We also give two examples from the new dataset for qualitative comparison in Fig.~\ref{fig:newdata}.

\section{Ablation Study}
In this section, we investigate the smoothing effect that is contributed by the two parts of the guidance. (No guidance, only texture guidance, only structure guidance, and double guidance).

\begin{figure*}[!t]
\centering

\subfigure[Training MSE loss]{\includegraphics[width=0.44\linewidth]{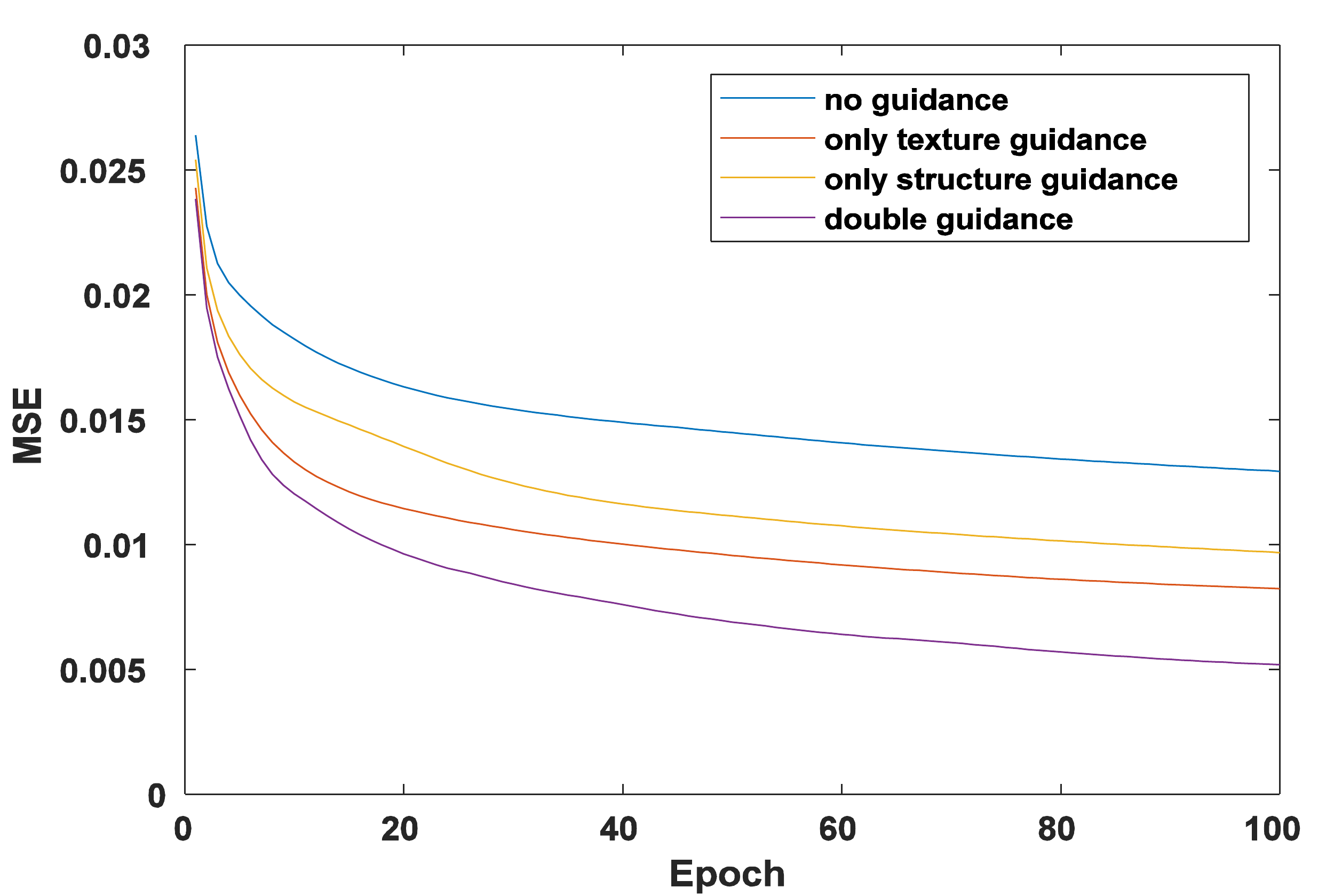}}
\hspace{10pt}
\subfigure[Validation MSE loss]{\includegraphics[width=0.44\linewidth]{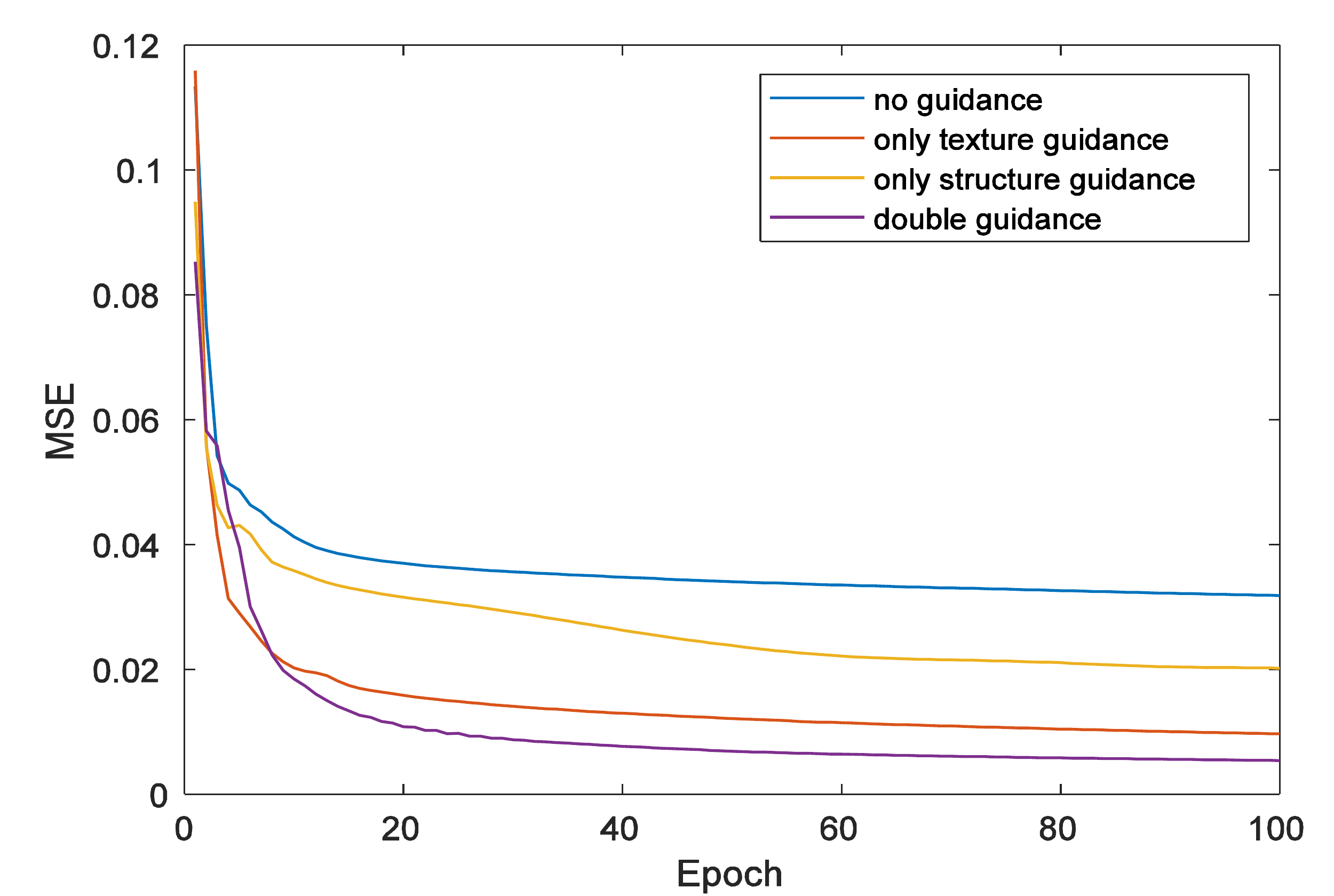}}

\caption{MSE loss of training and validation of four networks (without guidance, only texture guidance, only structure guidance, and double guidance). Overall, the loss with double guidance is the smallest in both training and validation process. It further indicates the effectiveness of using double guidance rather than single guidance or no guidance in image smoothing. 
}
\label{fig:mseloss}
\end{figure*}

\paragraph{\textbf{Training and validation loss}} We train the four networks (without guidance, only with structure guidance, only with texture guidance, and with double guidance) separately, and plot the MSE loss in 100 epochs in Fig.~\ref{fig:mseloss}. Compared with the results without guidance or with single guidance, the loss with double guidance is the smallest in both training and validation process. It can be seen that both parts of the guidance make an important contribution to overall performance. It further indicates the effectiveness of applying double guidance into image smoothing.

\begin{figure*}[!t]
\centering
\subfigure{\includegraphics[width=\linewidth]{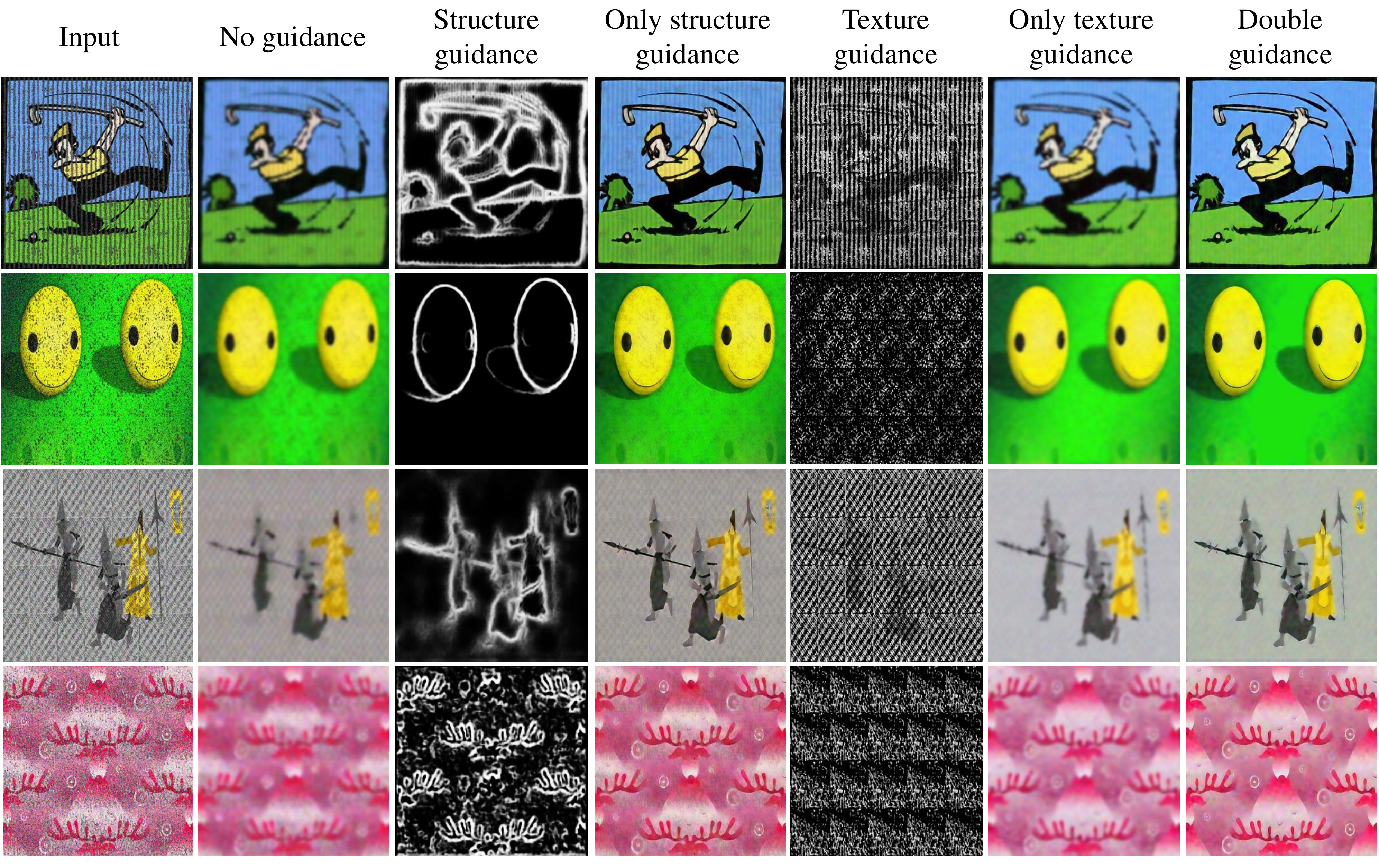}}\\
\caption{
Image smoothing results with different guidance on generated images.
}
\label{fig:synAblation}
\end{figure*}

\paragraph{\textbf{Qualitative comparison}}
We show several examples, including both generated (Fig.~\ref{fig:synAblation})  and natural images (Fig.~\ref{fig:naAblation}), to visually compare the smoothing results with different guidance. Overall, compared with the baseline (without any guidance), the results with only structure guidance can retain structures, as well as those of some strong textures. In contrast, the results with only texture guidance can smooth out textures, both strong and weak, more effectively. However, the main structures are obviously blurred. With double guidance, the filter takes advantage of the two properties and performs well in both preserving structures and removing textures.

\section{Comparison with Other Methods}
We visually compare our smoothing results with  Total Variation (TV) \cite{22rudin1992nonlinear}, L0 \cite{4xu2011image}, Relative Total Variation (RTV) \cite{23xu2012structure}, guided filter (GF) \cite{2he2013guided}, Structure Gradient and Texture Decorrelation (SGTD) \cite{9liu2013sgtd}, rolling guidance filter (RGF) \cite{11zhang2014rolling}, fast L0 \cite{26nguyen2015fast}, segment graph filter (SGF) \cite{3zhang2015segment}, static and dynamic filter (SDF) \cite{12ham2015robust}， and double-guided filter (DGF) \cite{53Lu_2017_DICTA}. We use the default parameters defined in their open-source code.

Fig.~\ref{fig:synComp} and Fig.~\ref{fig:naComp} show image smoothing results on our generated and natural images respectively. They show that our filter performs consistently well in both circumstances in terms of structure preservation and texture removal.

To investigate the image smoothing performance of different deep models \cite{46xu2015deep,48li2016deep,47liu2016learning,52Chen_2017_ICCV,49Fan_2017_ICCV}, we additionally give two challenging examples in Fig.~\ref{fig:deep}. Note that all the models are trained on our dataset. It turns out that our model can remove textures while preserving structures more effectively.

\section{Challenging Case}

We give a challenging case in Fig.~\ref{fig:challenge}, where the eyes, nose, and number of the runner are totally removed as textures. But actually, they have important semantic meaning in the real world. The HED we use for constructing structure guidance pays more attention to the object boundary, rather than details within the object, so it does not give reasonable confidence to these important details. Also, our texture prediction network cannot distinguish them as well. Thus, there is still a long way to go before achieving ideal smoothing results.

\section{Applications}
Image smoothing is a fundamental technology in image processing and computer vision with a broad range of applications. In the following, we mainly study three typical applications: image abstraction, detain enhancement, and edge detection.

\paragraph{\textbf{Image abstraction}}

Image abstraction aims to create a cartoon-like style from an input image. We use the method in \cite{30winnemoller2006real} for image abstraction, which involves smoothing the input and retaining main structures, detecting difference-of-Gaussian edges, and abstracting the image with soft color quantization. Fig.~\ref{fig:abs} lists four examples, where we study the abstraction results of the original input and the smoothed image respectively. Obviously, after smoothing, the abstraction results have less noise and artifacts. Further, the structures are sharpened, indicating the effectiveness of image smoothing.

\paragraph{\textbf{Detail enhancement}}
Suppose $I$ is the input image, and $S$ is the smoothed output. We define detail enhancement $DE$ as: $DE = S + \alpha  \cdot (I - S)$, where $\alpha  \ge 1$ controls the extent ($\alpha  = 2$ in this case). The results with different methods are shown in Fig.~\ref{fig:detail}. Our method is able to boost the details without affecting the overall color tone and without causing halos near structures.

\paragraph{\textbf{Edge detection}}
Image smoothing can also function as an essential pre-processing step in many other visual tasks, like edge detection. In Fig.~\ref{fig:edge}, we show the outcome of applying Canny edge detection to the original input and its smoothed version for the different guidance components. It is clear that with image smoothing, the Canny edges are clearer and more refined with less influence by insignificant details. We expect image smoothing to play a more significant role in other tasks. We will focus on this in future work.

\begin{figure*}[!t]
\centering
\subfigure{\includegraphics[width=\linewidth]{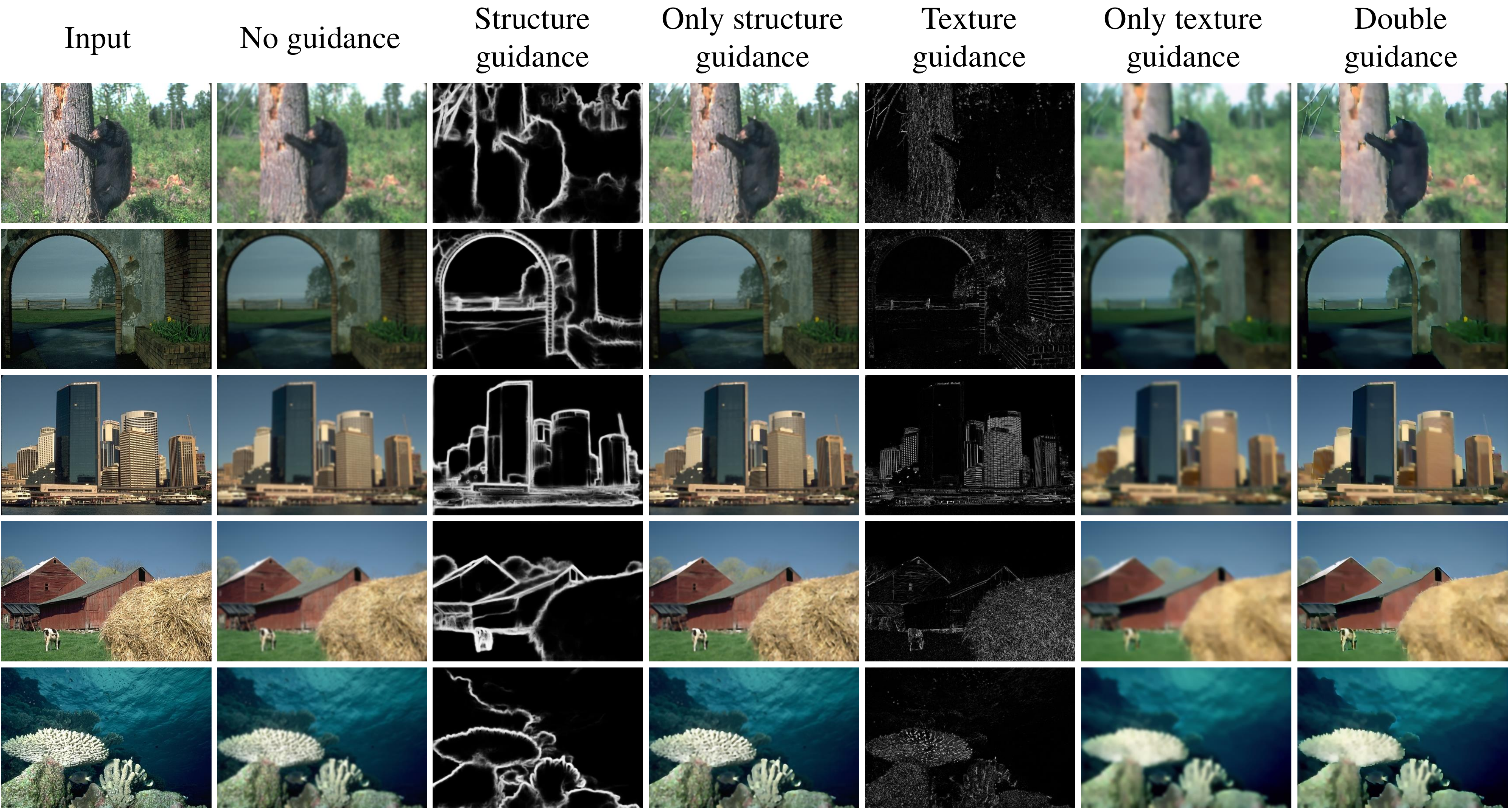}}\\
\label{fig:naAblation}
\end{figure*}
\begin{figure*}[!t]
\centering
\subfigure{\includegraphics[width=\linewidth]{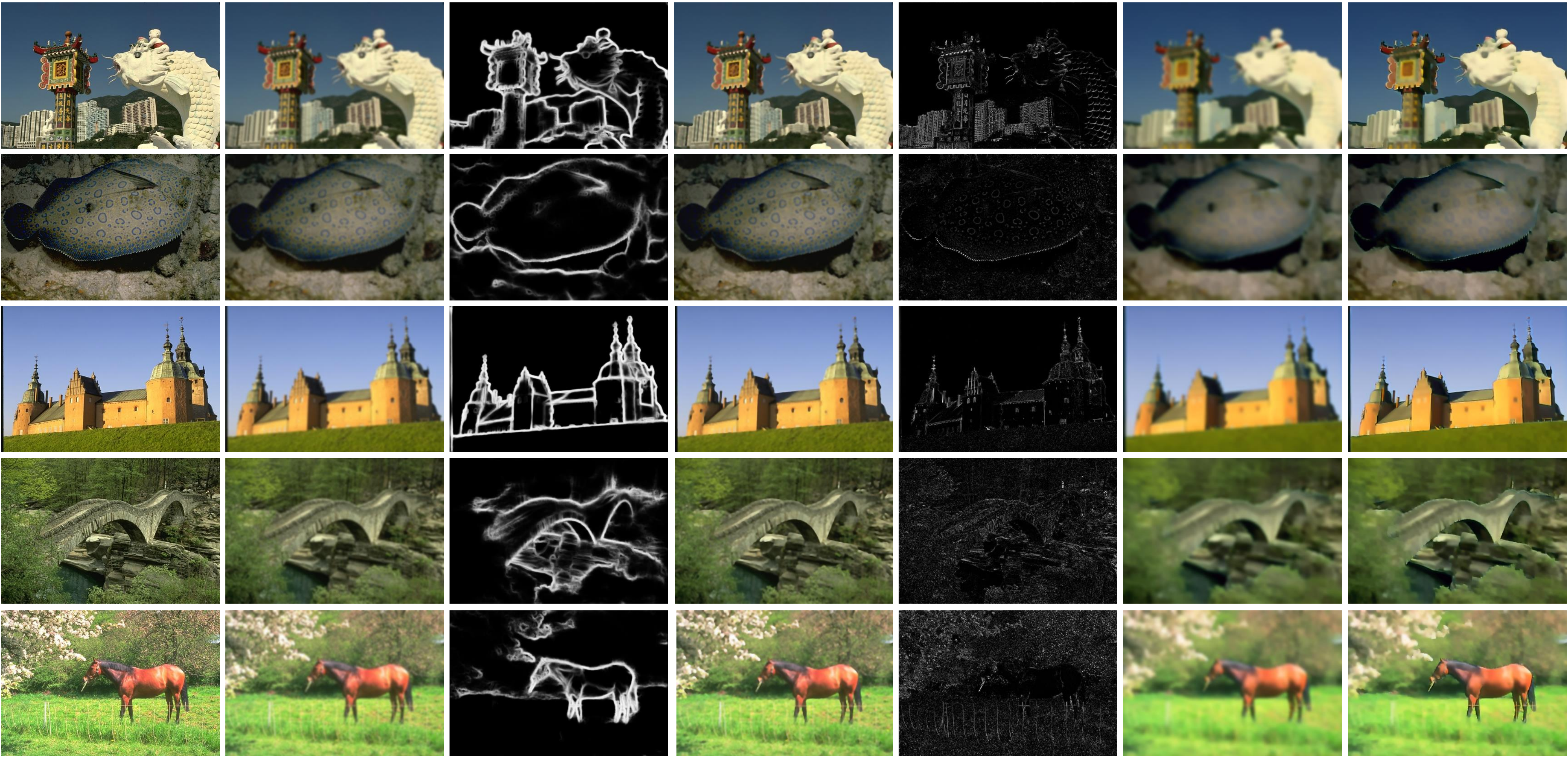}}\\
\caption{
Image smoothing results with different guidance on natural images.
}
\label{fig:naAblation}
\end{figure*}

\begin{figure*}[!t]
\centering
\subfigure{\includegraphics[width=\linewidth]{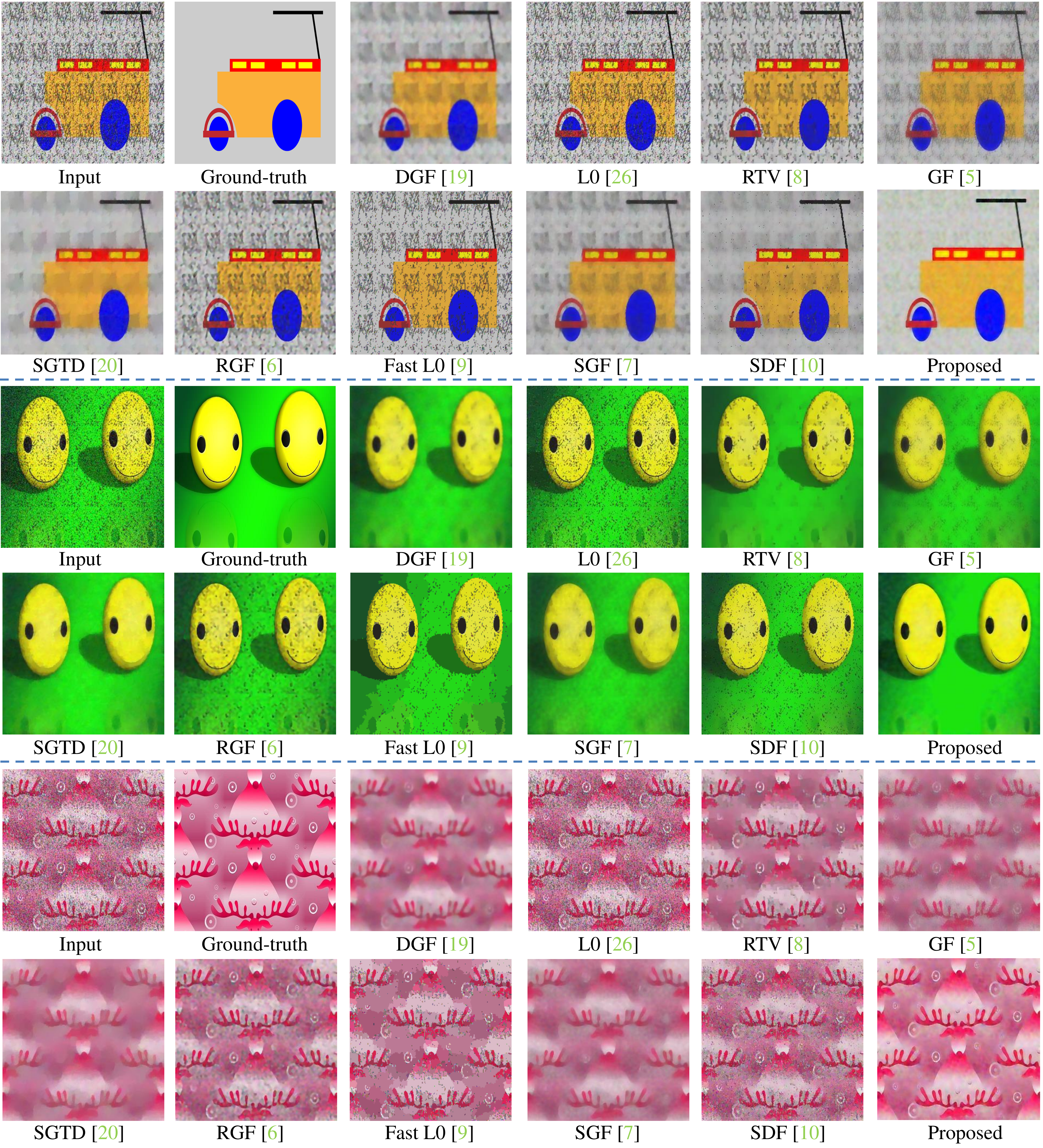}}\\
\caption{
Generated image smoothing results.
}
\label{fig:synComp}
\end{figure*}

\begin{figure*}[!t]
\centering
\subfigure{\includegraphics[width=0.945\linewidth]{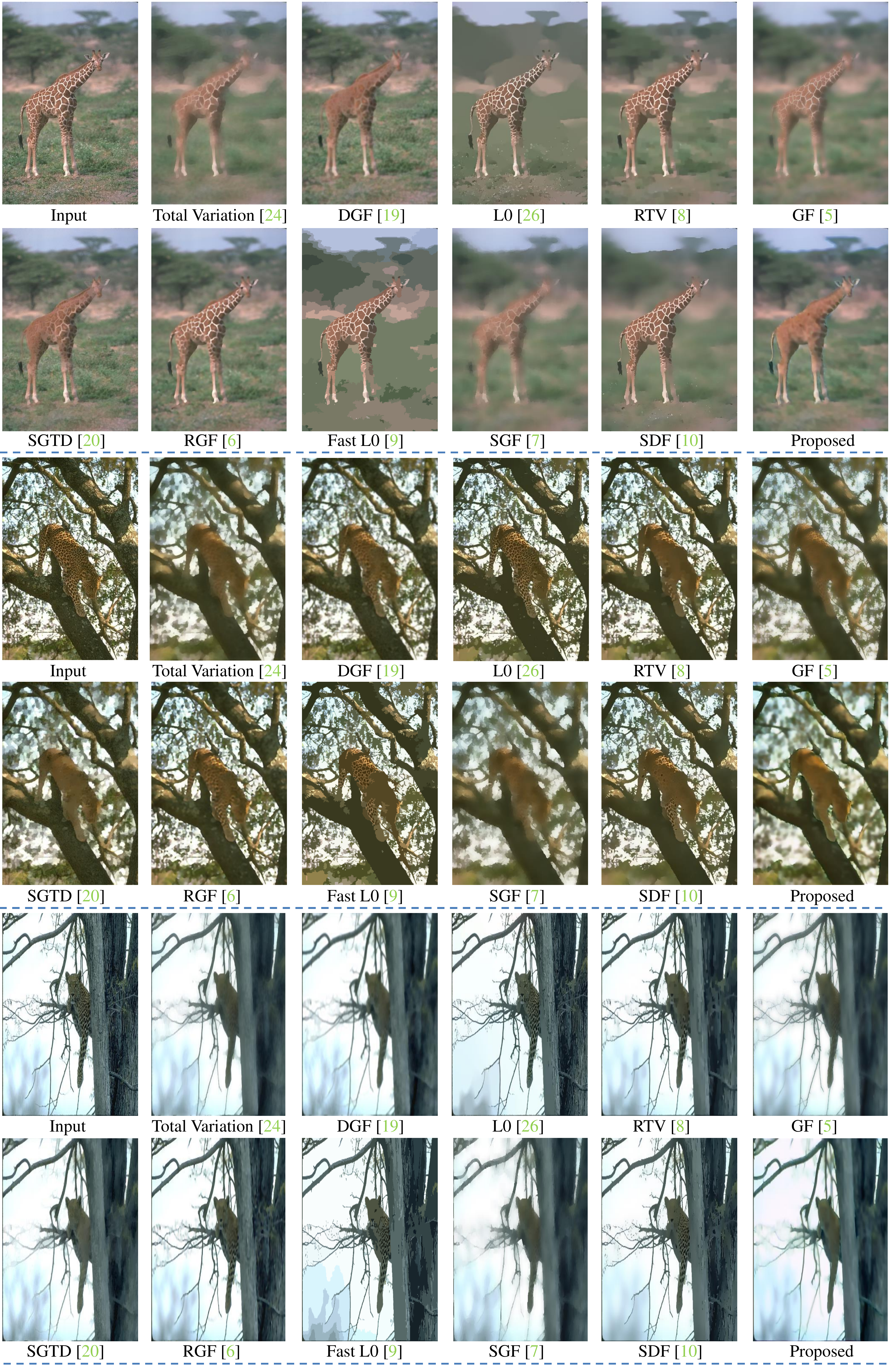}}\\
\label{fig:naComp}
\end{figure*}
\begin{figure*}[!t]
\centering
\subfigure{\includegraphics[width=\linewidth]{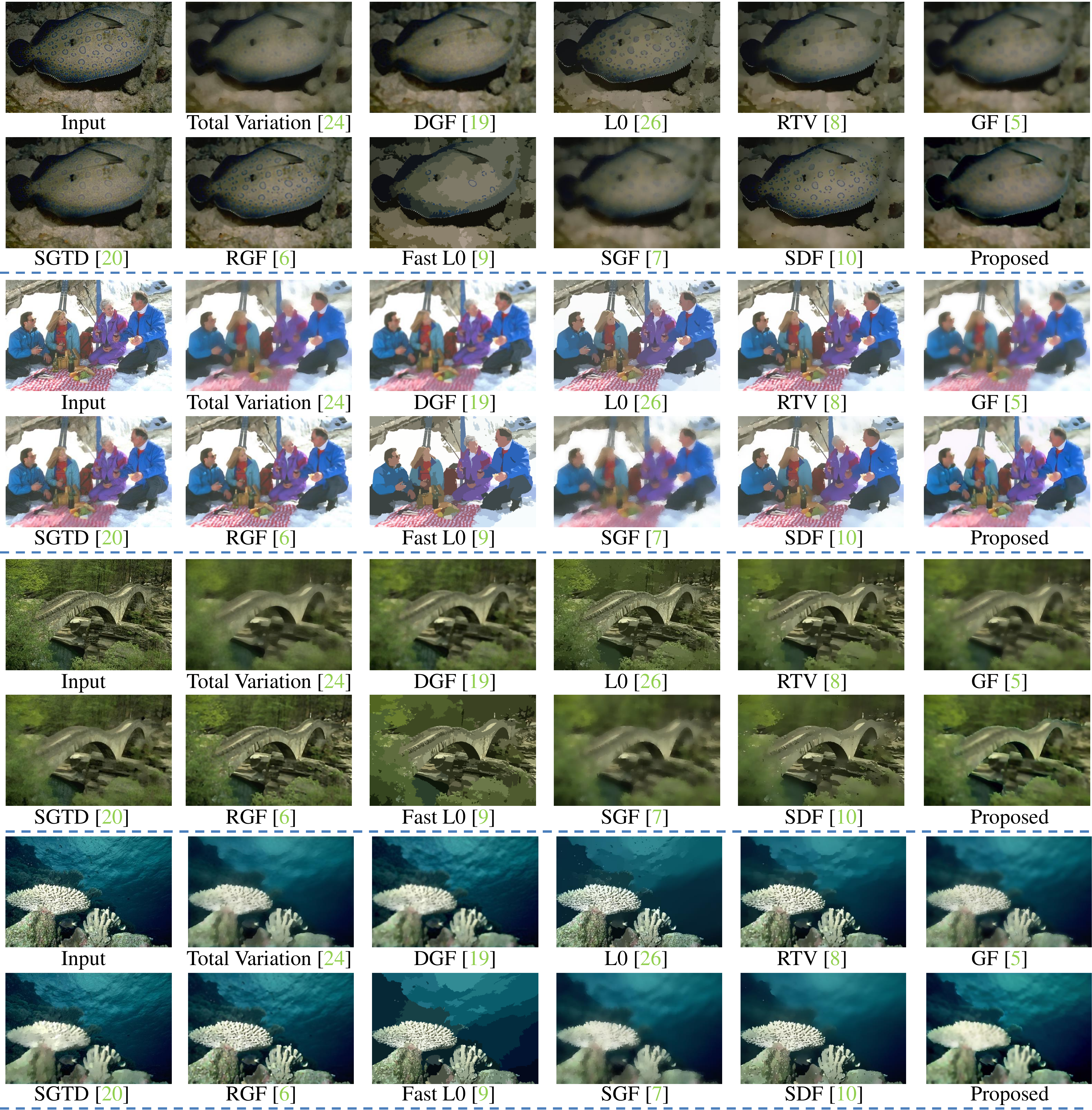}}\\
\label{fig:naComp}
\end{figure*}
\begin{figure*}[!t]
\centering
\subfigure{\includegraphics[width=\linewidth]{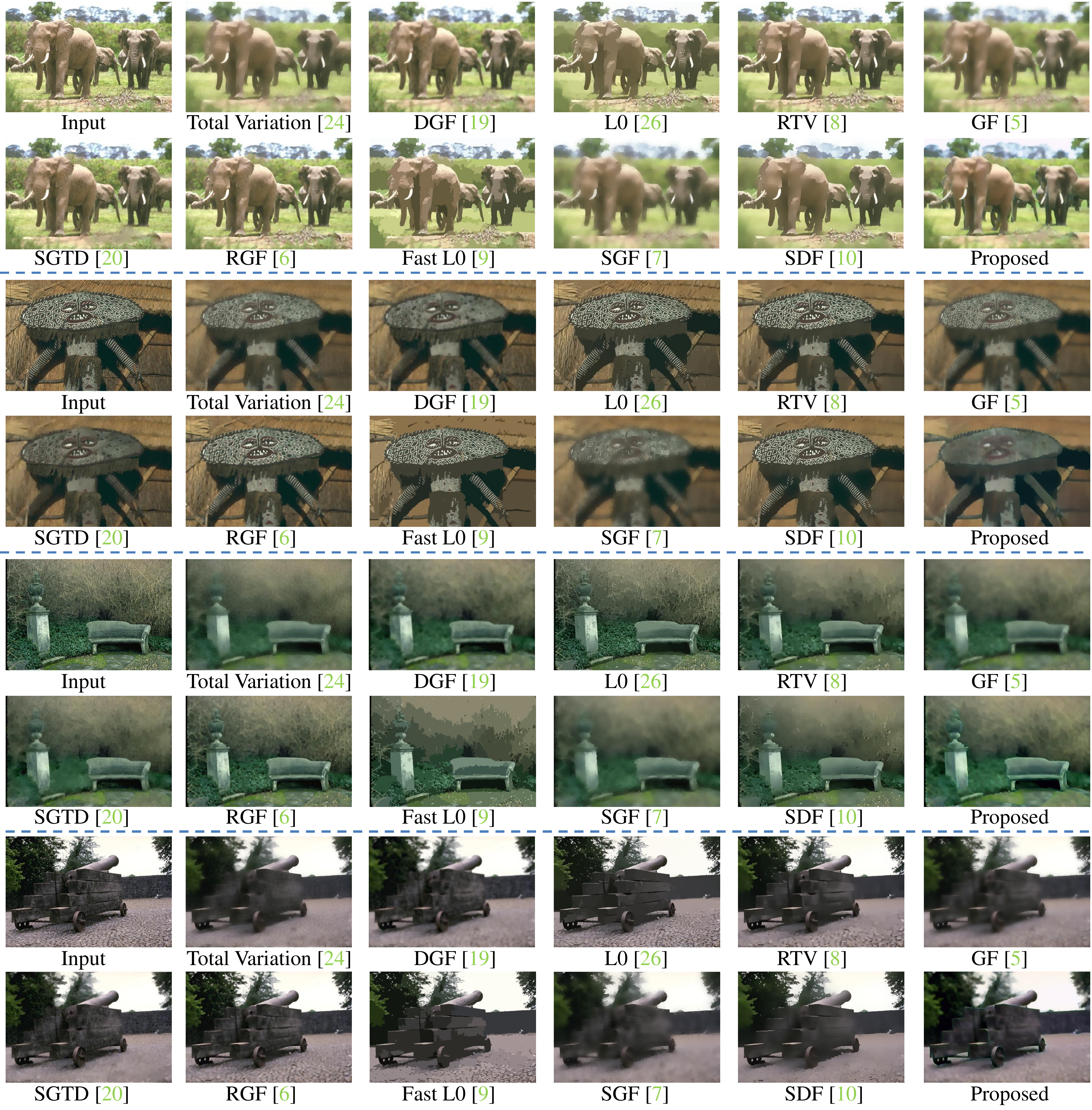}}\\
\caption{
Natural image smoothing results.
}
\label{fig:naComp}
\end{figure*}

\begin{figure*}[!t]
\centering
\subfigure{\includegraphics[width=\linewidth]{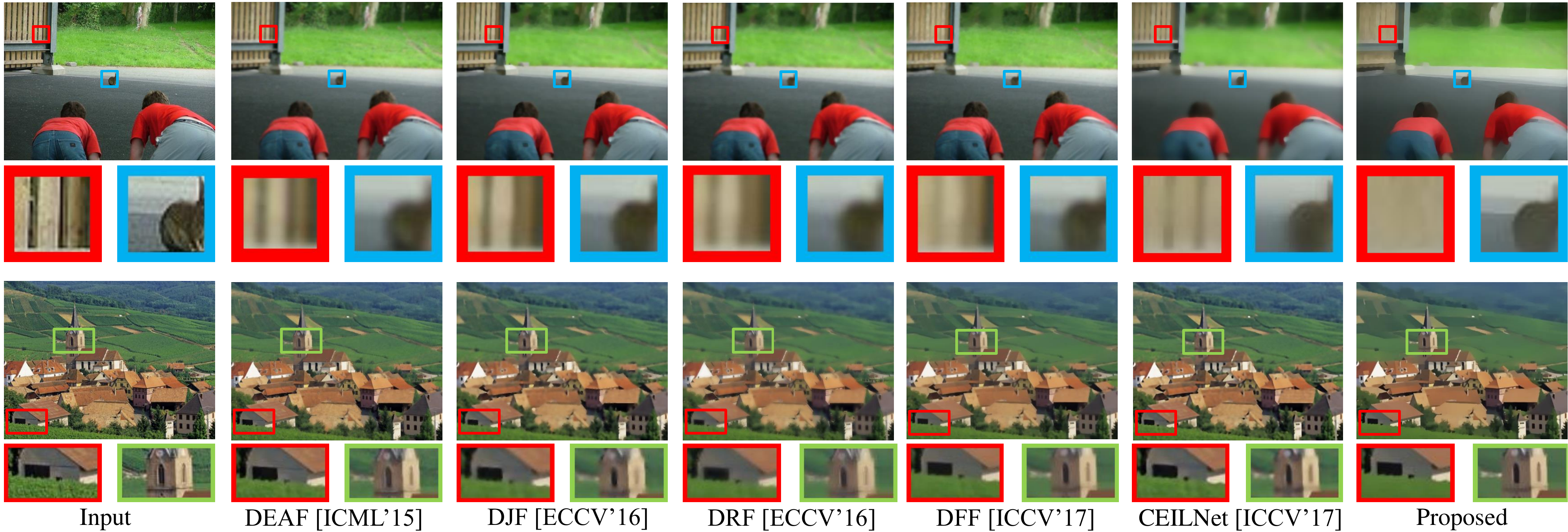}}\\
\caption{
Image smoothing results with different deep networks trained on our dataset. Our model performs better in removing textures and preserving structures at the same time.
}
\label{fig:deep}
\end{figure*}

\begin{figure*}[!t]
\centering
\subfigure{\includegraphics[width=\linewidth]{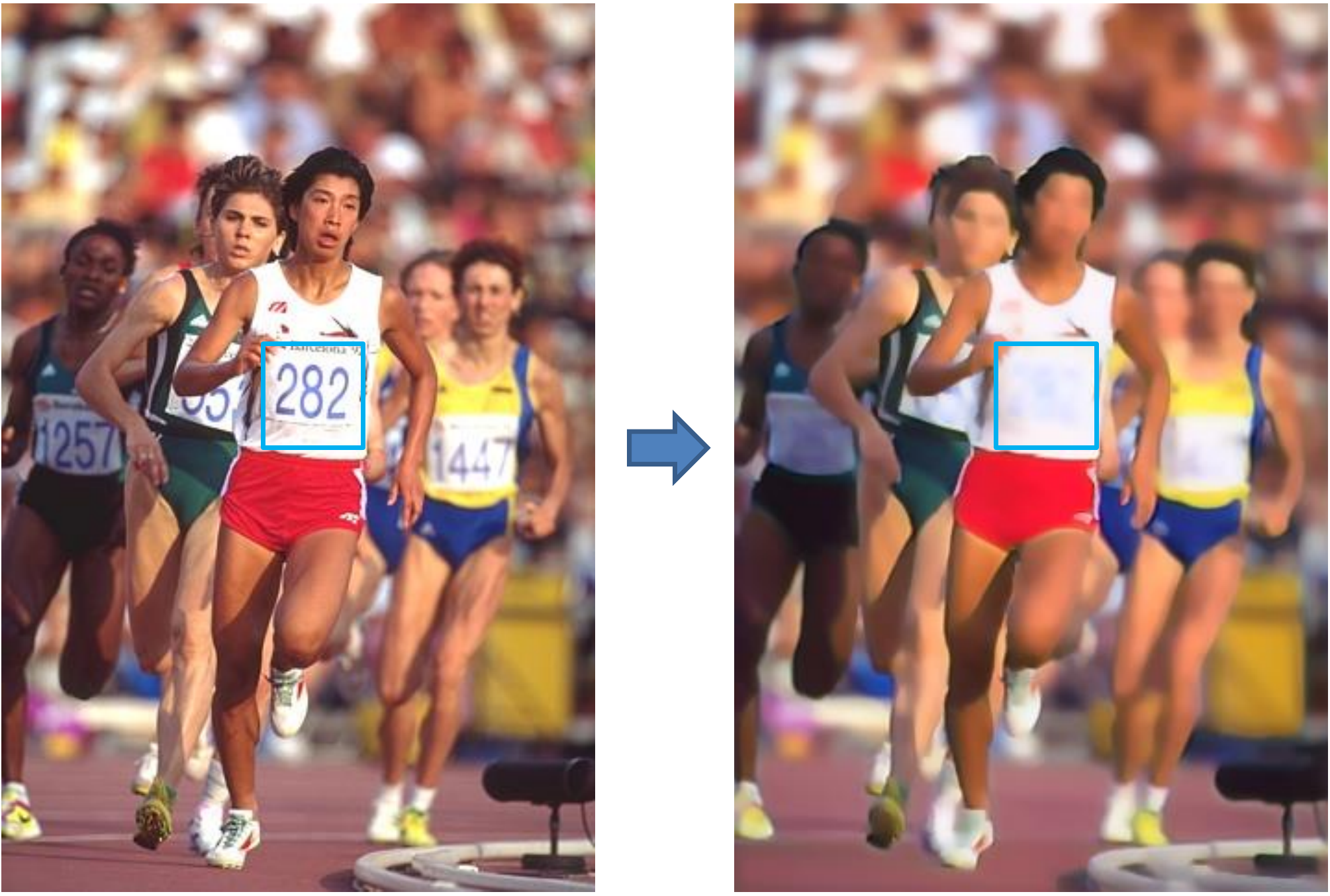}}\\
\caption{
Challenging case. The number, eyes, nose of the runner are smoothed out. Ideally, these should be preserved as they have significant semantic meaning.
}
\label{fig:challenge}
\end{figure*}

\begin{figure}[!t]
\centering
\subfigure{\includegraphics[width=0.9\linewidth]{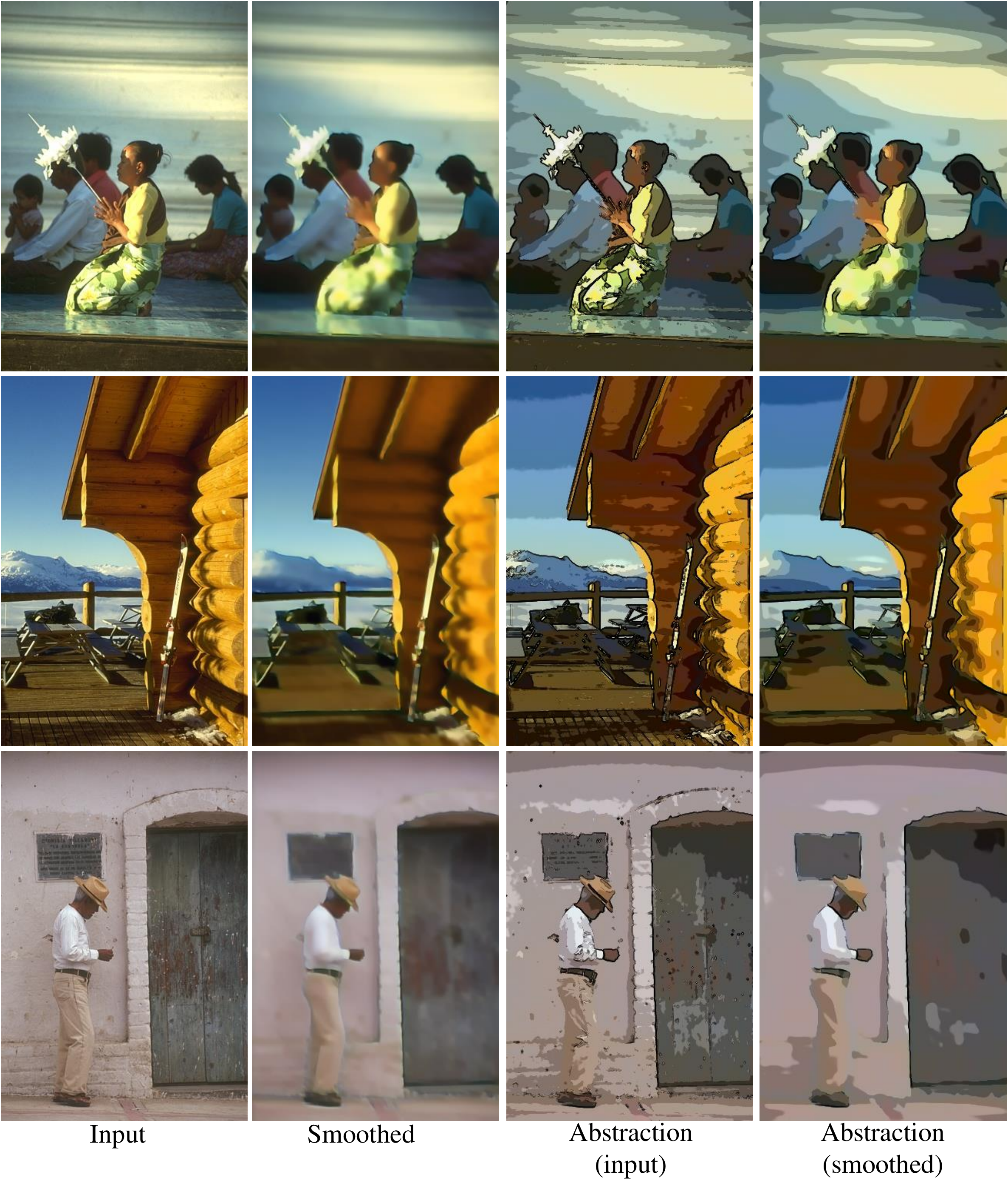}}\\
\caption{
Image abstraction results. Compared with the results to the original input directly, image smoothing can help to suppress more noise and artifacts, and sharpen the structures.
}
\label{fig:abs}
\end{figure}

\begin{figure}[!t]
\centering
\subfigure{\includegraphics[width=\linewidth]{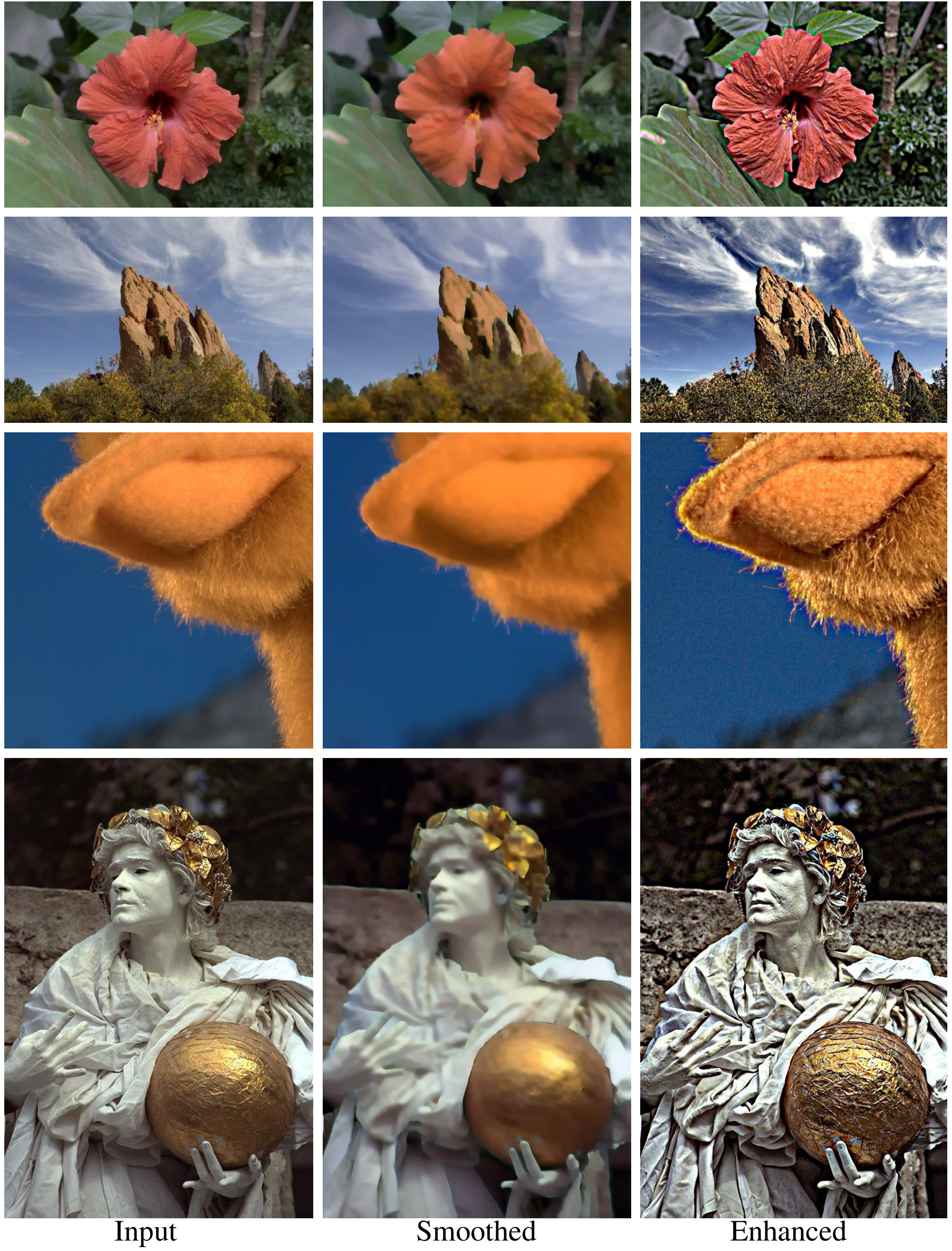}}\\
\caption{
Detail enhancement results. Our filter can boost the details without affecting the overall color tone and causing halos near structures.
}
\label{fig:detail}
\end{figure}

\begin{figure*}[!t]
\centering
\subfigure{\includegraphics[width=\linewidth]{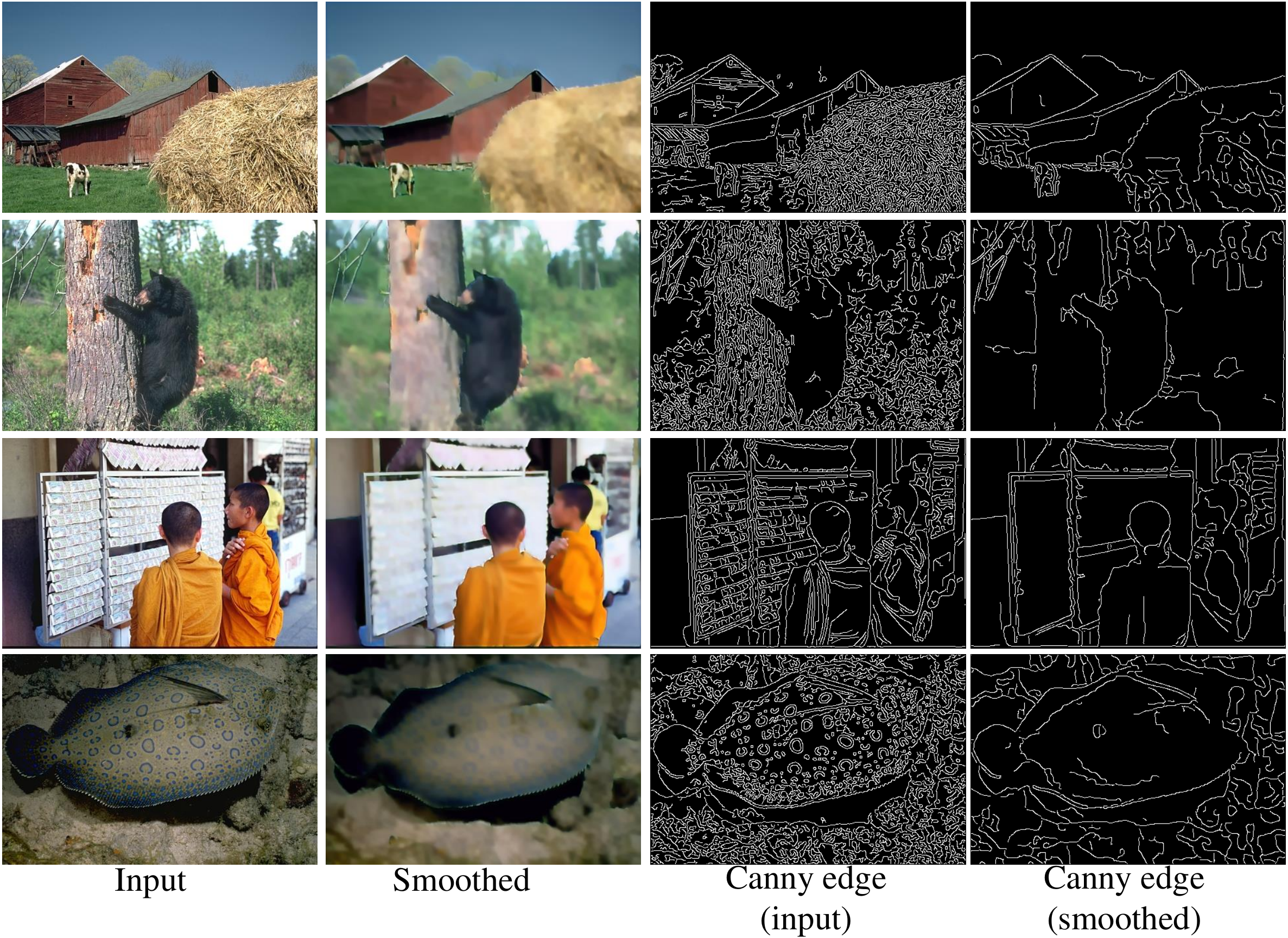}}\\
\caption{
Canny edge detection results. After smoothing, the edges are clearer and more refined with less influence by insignificant details.
}
\label{fig:edge}
\end{figure*}

\end{document}